\documentclass[graybox]{svmult}

\usepackage{mathptmx}       
\usepackage{helvet}         
\usepackage{courier}        
\usepackage{type1cm}        
\usepackage{makeidx}         
\usepackage{graphicx}        
\usepackage{multicol}        
\usepackage[bottom]{footmisc}
\usepackage{url}
\usepackage{multirow}

\usepackage{epstopdf}
\usepackage{multirow}
\usepackage{booktabs}
\usepackage{todonotes}
\usepackage{lipsum}
\usepackage{hyperref}
\hypersetup{hidelinks}

\usepackage{ifthen}
\usepackage{makeglos}

\graphicspath{{./}{./graphics/}{./graphics/eps/}{./graphics/pdf/}}

\usepackage{multirow}
\usepackage[official]{eurosym}
\usepackage{subfigure}

\newcommand{\buildbook}{false}

\makeindex             

\makeglossary

\begin{document}

%
%
%

\ifthenelse{\equal{true}{\buildbook}}{
\title{Review of Face Presentation Attack Detection Competitions}
}
{
\title*{Review of Face Presentation Attack Detection Competitions}
}

\def\ie{{\em i.e.}}
\def\eg{{\em e.g.}}
\def\etal{{\em et al.}}
\def\vs{\emph{vs.}}


\author{Zitong Yu, Jukka Komulainen, Xiaobai Li and Guoying Zhao}


\institute{Zitong Yu \at Center for Machine Vision and Signal Analysis, University of Oulu, Finland\\ \email{zitong.yu@oulu.fi}
\and Jukka Komulainen \at Visidon Ltd, Finland\\ \email{jukka.komulainen@kapsi.fi}
\and Xiaobai Li \at Center for Machine Vision and Signal Analysis, University of Oulu, Finland\\ \email{xiaobai.li@oulu.fi}
\and Guoying Zhao \at Center for Machine Vision and Signal Analysis, University of Oulu, Finland\\ \email{guoying.zhao@oulu.fi}}

%

%
\maketitle

%

\abstract*{
...
}

\abstract{
Face presentation attack detection (PAD) has received increasing attention ever since the vulnerabilities to spoofing have been widely recognized. The state of the art in unimodal and multi-modal face anti-spoofing has been assessed in eight international competitions organized in conjunction with major biometrics and computer vision conferences in 2011, 2013, 2017, 2019, 2020 and 2021, each introducing new challenges to the research community. In this chapter, we present the design and results of the five latest competitions from 2019 until 2021. The first two challenges aimed to evaluate the effectiveness of face PAD in multi-modal setup introducing near-infrared (NIR) and depth modalities in addition to colour camera data, while the latest three competitions focused on evaluating domain and attack type generalization abilities of face PAD algorithms operating on conventional colour images and videos. We also discuss the lessons learnt from the competitions and future challenges in the field in general.
}

\section{Introduction}
\label{sec:introduction}
Presentation attacks (PAs) \cite{ACER}, commonly referred to also as spoofing, pose a serious security issue to biometric systems in general but automatic face recognition (AFR) systems in particular are easy to be deceived e.g., using images of the targeted person published in the web or captured from distance. Many works, such as~\cite{mohammadi2018deeply}, have concluded that face recognition systems are vulnerable to sensor-level attacks launched with different presentation attack instruments (PAI), such as prints, displays and wearable 3D masks. The vulnerability to PAs is one of the main reasons to the lack of public confidence in AFR systems especially in high-security level applications, such as mobile payment services, which has created a necessity for robust solutions to counter spoofing.

One possible solution is to include a dedicated face presentation attack detection (PAD) component into an AFR system. Face PAD, commonly referred to also as face anti-spoofing (FAS) or liveness detection, aims at automatically differentiating whether the presented face sample originates from a bona fide subject or an artefact. Based on different kinds of camera sensors, face PAD schemes can be broadly categorized into unimodal and multi-modal based methods. Unimodal face PAD systems usually exploit efficient visual features for binary classification (\ie, bona fide vs. attack) classification extracted from conventional colour (RGB) camera data, thus can be easily deployed in most practical AFR scenarios but with limited accuracy. In contrast, with extra hardware costs, multi-modal methods~\cite{wan2020multi} introduce some additional imaging modalities (\eg, depth, near-infrared (NIR), or thermal infrared sensor data) that can capture specific intrinsic differences between the bona fide and attack samples. For example, the depth maps obtained from 2D printout and display facial artefacts using 3D sensors usually have flat and close-to-zero distributions in facial regions.

Despite the recent progress in deep learning based face anti-spoofing methods~\cite{ming2020survey,yu2021deep} with powerful representation capacity, it is difficult to tell what are the best or most promising feature learning approaches for generalized face PAD. Along with the development in manufacturing technologies, it has become even cheaper for an attacker to exploit a known vulnerability of a face authentication system with different kinds of facial artefacts, such as a realistic 3D mask made of plaster. Simulating AFR scenarios with various attacks, environmental conditions, acquisition devices and subjects is extremely time-consuming and expensive but the domain shifts caused by such covariates have a significant impact on the face PAD performance. These issues have been already explored with several public unimodal datasets, such as~\cite{Boulkenafet2017OULU,Chingovska_BIOSIG,Liu2018Learning,Wen2015Face,Zhang2012A}, but these benchmarks have been yet rather small-scale in terms of number of subjects, samples and AFR scenarios and, consequently, the corresponding evaluation protocols have been too limited (\eg, in terms of unknown PAs and acquisition conditions in the test set). Moreover, there have been no public benchmark and protocols for evaluating multi-modal face PAD schemes.

Competitions play a key role in advancing the research on face PAD and provide valuable insights for the entire face recognition community. It is important to organize collective evaluations regularly in order to assess, or ascertain, the current state of the art and gain insight on the robustness of different approaches using a common platform. Also, new more challenging public datasets are often collected and introduced within such collective efforts to the research community for future development and benchmarking use. The quality of PAIs keeps improving as technology (e.g., 2D/3D printers and displays) gets cheaper and better, which is another reason why benchmark datasets need to be updated regularly. Open contests are likely to inspire researchers and engineers beyond the field to participate, and their outside the box thinking may lead to new ideas on the problem of face PAD and novel countermeasures.

In the context of face PAD, altogether eight international competitions~\cite{ChallengeCVPR2019,boulkenafet2017competition,chakka2011competition,chingovska20132nd,cvpr2020challenge,liu20213d,purnapatra2021face,zhang2021celeba} have been organized in conjunction with major biometrics and computer vision conferences in 2011, 2013, 2017, 2019, 2020 and 2021, each introducing new challenges to the research community. In this chapter, we focus on analysing the design and results of the five latest competitions~\cite{ChallengeCVPR2019,cvpr2020challenge,liu20213d,purnapatra2021face,zhang2021celeba} from 2019 until 2021, while an extensive review of the first three competitions~\cite{boulkenafet2017competition,chakka2011competition,chingovska20132nd} can be found in \cite{komulainen2019review}. The key features of the five most recent face PAD competitions are summarized in Table~\ref{tab:introduction}. 

\begin{table*}[h]
\caption{Summary of the recent five face PAD competitions organized from 2019 until 2021. \label{tab:introduction}}
{
\scalebox{0.75}{
\begin{tabular}{c|c|c|c}
\toprule
Competition   &  Modality   & Highlight  & Limitation   \\ \midrule

CVPR2019 challenge~\cite{ChallengeCVPR2019}
& RGB, depth, NIR
& First multi-modal face PAD challenge &  With only print attacks \\ 

CVPR2020 challenge~\cite{cvpr2020challenge}
& RGB, depth, NIR
& Cross-ethnicity \& cross-PAI testing  &  Testing with only print and mask PAs   \\ 
 
ECCV2020 challenge~\cite{zhang2021celeba}
& RGB
& Largest dataset with rich (43) attributes  &  Limited domain shift in testing set  \\ 

IJCB2021 LivDet-Face~\cite{purnapatra2021face}
& RGB
& Unseen testing set with rich (9) PAIs &  No training set provided \\ 

ICCV2021 challenge~\cite{liu20213d}
& RGB
& Largest 3D mask dataset \& open set protocol  &  Limited (only three) mask types                \\ \bottomrule
\end{tabular}
}
}{}
\end{table*}

The multi-modal face anti-spoofing challenge organized in 2019 (referred to as CVPR2019 challenge)~\cite{ChallengeCVPR2019} provided an initial assessment of multi-modal countermeasures to various kinds of print attacks by introducing a precisely defined test protocol for evaluating the performance of the face PAD solutions with three modalities (\ie, colour, depth and NIR). In 2020, the cross-ethnicity face anti-spoofing challenge (referred to as CVPR2020 challenge)~\cite{cvpr2020challenge} extended the previous CVPR2019 challenge with several new factors (\eg, unseen ethnicities and 3D mask attacks), and having separate competition tracks for unimodal (colour) and multi-modal (colour, depth and NIR) data. While the datasets used in the first two contests contained a limited number of subjects and samples, the CelebA-Spoof Challenge in 2020 (referred to as ECCV2020 challenge)~\cite{zhang2021celeba} provided an assessment on the performance of face PAD methods on the largest publicly available unimodal (colour) benchmark dataset. In order to bridge the gap between competitions and real-world application scenarios, in the LivDet Face 2021 liveness detection competition (referred to as IJCB2021 LivDet-Face)~\cite{purnapatra2021face}, the test data was practically concealed as only few samples of attacks were provided, thus the evaluation focused on the challenging domain generalization issues and unseen attacks with separate competition tracks for methods using colour image and video data. Finally, to further evaluate the performance of face PAD approaches under challenging 3D mask attacks, the 3D high-fidelity mask attack detection challenge in 2021 (referred to as ICCV2021 challenge)~\cite{liu20213d} was conducted on the largest publicly available 3D mask dataset with a novel open-set evaluation protocol.

The remainder of the chapter is organised as follows. First, we will recapitulate the organization, solutions as well as results of the five most recent face PAD competitions in Section~\ref{sec:competitions}. Then, in Section~\ref{sec:discussion}, we will discuss the lessons learnt from the competitions and future challenges in the field of face PAD in general. Finally, Section~\ref{sec:conclusion} summarizes the chapter, and presents conclusions drawn from the competitions discussed here.

\section{Review on recent face PAD competitions} 
\label{sec:competitions}

We begin our review by first introducing two multi-modal face PAD competitions, namely CVPR2019 and CVPR2020 challenges, in Sections~\ref{sec:com1} and~\ref{sec:com2}, respectively, where the latter one included also a competition track for unimodal (colour) data. Then, three latest unimodal (colour) face PAD competitions, namely ECCV2020, IJCB2021 LivDet-Face and ICCV2021 challenges are reviewed in remaining Sections~\ref{sec:com3},~\ref{sec:com4} and~\ref{sec:com5}, respectively.

\subsection{Multi-modal Face Anti-spoofing Attack Detection Challenge (CVPR2019)}
\label{sec:com1}


The first three face PAD competitions~\cite{komulainen2019review} organized in conjunction of International Joint Conference on Biometrics (IJCB) 2011, International Conference on Biometrics (ICB) 2013 and IJCB 2017 focused on photo (\ie, both printed and digital) and video-replay attack detection relying on small-scale datasets (\ie, PRINT-ATTACK~\cite{chakka2011competition}, REPLAY-ATTACK~\cite{chingovska20132nd}, OULU-NPU~\cite{boulkenafet2017competition}) for training, tuning and testing. To be more specific, these datasets have insufficient number of subjects ($\textless 60$) and data samples ($\textless 6,000$ videos) compared with databases used in the field of image classification, \eg, ImageNet~\cite{deng2009imagenet} or face recognition, \eg, CASIA-WebFace~\cite{yi2014learning}, which severely limits the development and testing of data-driven deep model based approaches for generalized face PAD. Also, due to the lack of variation in the face PAD datasets, the deep models have been suffering from overfitting and learning database-specific information instead of generalized feature representations capturing the disparities in the inherent fidelity characteristics between bona fide samples and different kinds of facial artefacts. Another missing feature in previous face PAD competitions has been the availability of multi-modal facial information in addition to conventional visible light colour (RGB) data. This kind of extended range image information might be very helpful for developing more robust face PAD methods for practical real-world AFR applications. In order to address the limitations of previous competitions, the Chalearn multi-modal face anti-spoofing attack detection challenge\footnote{\url{https://sites.google.com/qq.com/face-anti-spoofing/welcome/challengecvpr2019}}~\cite{liu2019multi} was held in conjunction with the Conference on Computer Vision and Pattern Recognition (CVPR) 2019. The competition was based on a new collected large-scale multi-modal face anti-spoofing dataset, namely CASIA-SURF~\cite{zhang2020casia,zhang2019dataset}, which consists of $1,000$ subjects and $21,000$ video clips with three modalities (colour, depth and NIR). The goal of this competition was to push the research progress in the AFR applications, where plenty of data and multiple modalities can be considered to be available.

\begin{table}[]
	\centering
	\caption{Teams and affiliations listed in the final ranking of the CVPR2019 challenge~\cite{liu2019multi}.}
	\begin{tabular}{|c|c|c|}
		\hline
		Ranking & Team Name        & Affiliation                                  \\ \hline
		\hline
		1 & VisionLabs       & VisionLabs                                         \\ \hline
		2 & ReadSense        & ReadSense                                          \\ \hline
		3 & Feather          & Intel                                              \\ \hline
		4 & Hahahaha         & Megvii                                             \\ \hline
		5 & MAC-adv-group    & Xiamen University                                  \\ \hline
		6 & ZKBH             & Biomhope                                           \\ \hline
		7 & VisionMiracle    & VisonMarcle                                        \\ \hline
		8 & GradiantResearch & Gradiant                                           \\ \hline
		9 & Vipl-bpoic       & ICT, CAS                                           \\ \hline
		10 & Massyhnu         & Hunan University                                  \\ \hline
		11 & AI4all           & BUPT                                              \\ \hline
		12 & Guillaume        & Idiap Research Institute                          \\ \hline
		invited team   & Vivi             & Baidu                                 \\ \hline
	\end{tabular}
	\label{table:team1}
	
\end{table}

The CVPR2019 challenge was run in the CodaLab\footnote{\url{https://competitions.codalab.org/competitions/20853}} platform and consisted of two phases: development phase (Dec. 22, 2018 - March 6, 2019) and final phase (March 6, 2019 - March 10, 2019). More than 300 academic research and industrial institutions worldwide participated in this challenge, and finally 13 teams entered into the final stage. A summary with the names and affiliations of these teams is presented in Table~\ref{table:team1}. Compared with the previous competitions~\cite{boulkenafet2017competition,chakka2011competition,chingovska20132nd}, the majority of the final participants (10 out of 13) of this competition came from the industry, which indicates the increased need for realiable liveness detection products in daily life applications. Furthermore, one highlight of the CVPR2019 challenge is that the three top-performing teams (VisionLabs\footnote{\url{https://github.com/AlexanderParkin/ChaLearn_liveness_challenge}}, ReadSense\footnote{\url{https://github.com/SeuTao/CVPR19-Face-Anti-spoofing}}, and Feather\footnote{\url{https://github.com/SoftwareGift/FeatherNets\_Face-Anti-spoofing-Attack-Detection-Challenge-CVPR2019}}) released their source code in GitHub and summarized their approaches in the related CVPR workshop papers~\cite{parkin2019recognizing,shen2019facebagnet,wang2019multi,zhang2019feathernets}, enhancing the fairness, transparency, and reproducibility of the solutions so that they can be easily facilitated by the face recognition community.

\begin{figure}[t]
	\begin{center}
	\includegraphics[width=1.0\linewidth]{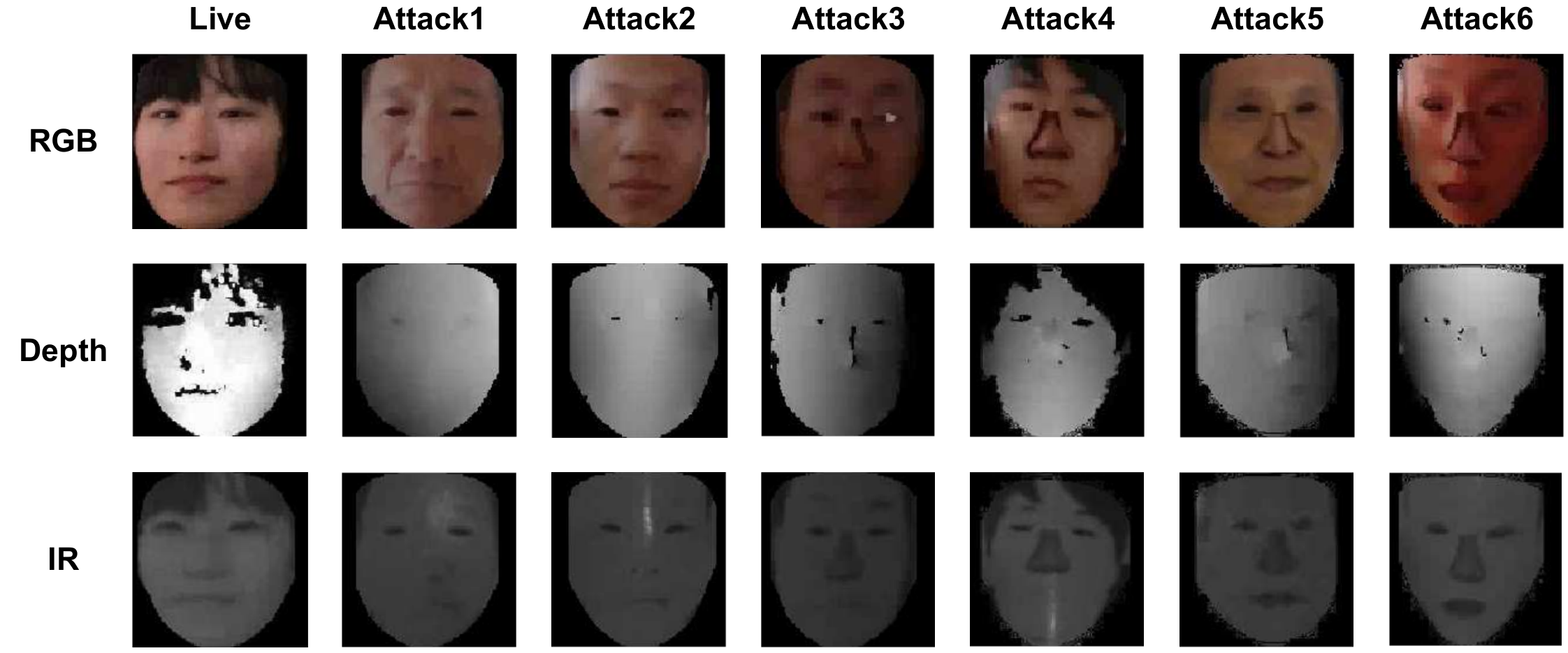}
	\end{center}
	\caption{Samples of a live face and six kinds of print attacks from the CASIA-SURF dataset~\cite{zhang2019dataset,zhang2020casia}.}
	\label{fig:CASIASURF}
\end{figure}

\subsubsection{Dataset}
\label{CASIA-SURF_dataset}


The CASIA-SURF dataset~\cite{zhang2019dataset,zhang2020casia} was the largest face PAD dataset in terms of number of subjects and videos at the time of the CVPR19 challenge. Each sample of the dataset was associated with three modalities (colour, depth and NIR) captured with an Intel RealSense SR300 camera. Samples from each subject consist of one live video clip and one video clip of each six different attack presentations. A total number of $1,000$ subjects and $21,000$ videos were captured for building the dataset. Representative samples of bona fide and attack samples across the three modalities are illustrated in Fig.~\ref{fig:CASIASURF}. 

The CASIA-SURF dataset considers six different kinds of print attacks, where a person is holding:

$\bullet$ \textbf{Attack 1:} A flat face photo from which the eye regions are cut .

$\bullet$ \textbf{Attack 2:} A curved face photo from which the eye regions are cut.

$\bullet$ \textbf{Attack 3:} A flat face photo from which the eye and nose regions are cut.

$\bullet$ \textbf{Attack 4:} A curved face photo from which the eye and nose regions are cut.

$\bullet$ \textbf{Attack 5:} A flat face photo from which eyes, nose and mouth regions are cut.

$\bullet$ \textbf{Attack 6:} A curved face photo from which the eye, nose and mouth regions have been cut.

The samples in the CASIA-SURF dataset were pre-processed as follows for the competition: 1) The dataset is split in three subject-disjoint partitions: train, validation and test sets, with 300, 100 and 600, respectively, when the corresponding number of videos is 6,300 (2,100 per modality), 2,100 (700 per modality) and 12,600 (4,200 per modality). 2) Only every tenth frame from each video was selected to reduce the size of the competition dataset, which resulted in 148K, 48K and 295K frames for the three subsets, respectively. 3) To mitigate the effect of pre-processing methods (\eg, face detection and alignment) and limit the problem of face PAD to the actual facial information, the background information was masked out pixel-wise from original data, thus only pre-cropped aligned facial images were provided for each modality.

\subsubsection{Evaluation Protocol and Metrics}
\label{sec:metrics}

In order to focus on the generalization to unknown attacks, the organizers provided only a part of the CASIA-SURF dataset for training, \ie, for each subject only a subset of PA types was available. Hence, challenge participants were given about 30K frames for training and 9.6K frames for validation. Note that attacks in the test set differ from attacks in the training set, therefore a successful model should avoid intra-attack overfitting, which was a common issue in the earlier face PAD competitions. The challenge comprised development and final stages. The detailed protocols are described as follows.

\textbf{Protocol in development phase:}~(\emph{Dec. 22, 2018 - March 6, 2019}). During the development phase participants had access to the labeled training and unlabeled validation samples. Training data included the bona fide samples and three kinds of PAs (4, 5, 6), whereas the validation data consisted of bona fide samples and three other types of PAs (1, 2, 3). Participants were able to submit predictions on the validation partition and receive immediate response via the leaderboard using the CodaLab platform. As it can be observed from Fig.~\ref{fig:CASIASURF}, the attacks (4, 5, 6) in the validation set differ in appearance (partial cuts in eyes, nose, and mouth regions) from attacks (1, 2, 3), which made the face PAD challenging.

\textbf{Protocol in final phase:}~(\emph{March 6, 2019 - March 10, 2019}). During the final phase, the labels for the validation subset were made available to participants, so that they could leverage the additional labeled data for tuning to alleviate the domain gap between different attack types. The participants had to make predictions on the unlabeled test partition and upload their solutions to the CodaLab platform. The considered test set was formed from bona fide samples and three kinds of PAs (1, 2, 3). The final ranking of the participants was determined based on the performance of the submissions on the test set. To be eligible for prizes, the top solutions had to release their source code under a license of their choice and provide a fact sheet describing their solution, which would be checked and reproduced for the sake of fairness and reproducibility.

\textbf{Evaluation metrics:} The recently standardized ISO/IEC 30107-3\footnote{\url{https://www.iso.org/standard/67381.html}}~\cite{ACER} metrics, including Attack Presentation Classification Error Rate (APCER), Normal Presentation Classification Error Rate (NPCER) and Average Classification Error Rate (ACER) were adopted as part of the used evaluation metrics. They could be formulated as: \begin{equation}APCER=FP/\left(FP+TN\right ),\end{equation} \begin{equation}NPCER=FN/\left (FN+TP\right ),\end{equation} \begin{equation}ACER=\left ( APCER+NPCER \right )/2,\end{equation}
where TP, FP, TN and FN correspond to true positive, false positive, true negative and false negative, respectively. APCER and NPCER are used to measure the error rates of attack or bona fide samples, respectively. Similarly to the common metrics in AFR systems, the Receiver Operating Characteristic (ROC) curve was also considered for examining a suitable operating point trade-off in the false positive rate (FPR) and true positive rate (TPR) regarding the requirements of real-world biometric applications. Finally, the operating point of TPR@FPR=$10^{-4}$ was selected as the leading evaluation measure for the CVPR2019 challenge, while the ACER was used as an additional evaluation criterion.

\subsubsection{Results and Discussion}

In this subsection, we summarize all the face PAD solutions reaching the final stage in terms of method keywords, backbone models, pretraining data, modalities, fusion schemes and loss functions. Finally, the overall results are analyzed and discussed.

\begin{table*}[]
	\centering
	\scriptsize
	
	\caption{Summary of the face PAD methods for all participating teams and baseline method~\cite{liu2019multi}. 'SE' denotes Squeeze and Excitation~\cite{he2016deep} and  'BCE' denotes binary cross-entropy.}
	
	\begin{tabular}{|c|c|c|c|c|c|c|c|}
		\hline
		Team  & Method   & Backbone model  & Pre-training data  & Modalities    & \begin{tabular}[c]{@{}c@{}}Fusion scheme \\ and loss function\end{tabular} \\ \hline
		\hline
		VisionLabs~\cite{parkin2019recognizing}   & \begin{tabular}[c]{@{}c@{}}Fine-tuning\\ Ensembling\end{tabular}  & \begin{tabular}[c]{@{}c@{}}Resnet-34~\cite{he2016deep}\\ Resnet-50~\cite{he2016deep}\end{tabular}  & \begin{tabular}[c]{@{}c@{}}CASIA-WebFace~\cite{yi2014learning}\\ AFAD-Lite~\cite{niu2016ordinal}\\ MSCeleb1M~\cite{guo2016ms}\\ Asian dataset~\cite{zhao2018towards}\end{tabular} & \begin{tabular}[c]{@{}c@{}}RBG\\ Depth\\ NIR\end{tabular} &  \begin{tabular}[c]{@{}c@{}}SE Fusion\\ Score fusion\\ BCE loss\end{tabular} \\ \hline
		
		ReadSense~\cite{shen2019facebagnet}   & \begin{tabular}[c]{@{}c@{}}Bag-of-local\\ features\\ Ensembling\end{tabular}  & SEresnext~\cite{xie2017aggregated}   & No   & \begin{tabular}[c]{@{}c@{}}RBG\\ Depth\\ NIR\end{tabular} & \begin{tabular}[c]{@{}c@{}}SE Fusion\\ Score fusion\\ BCE loss\end{tabular} \\ \hline
		
		Feather~\cite{zhang2019feathernets}     & Ensembling    & \begin{tabular}[c]{@{}c@{}}Fishnet~\cite{sun2018fishnet} \\ MobileNetV2~\cite{sandler2018mobilenetv2}\end{tabular} & No    & \begin{tabular}[c]{@{}c@{}}Depth\\ NIR\end{tabular}     & \begin{tabular}[c]{@{}c@{}}Score fusion\\ BCE loss\end{tabular} \\ \hline
		
		Hahahaha  & \begin{tabular}[c]{@{}c@{}}Depth only\end{tabular}  & Resnext~\cite{xie2017aggregated}   & Imagenet~\cite{deng2009imagenet}   & Depth & BCE loss      \\ \hline

		MAC-adv-group   & \begin{tabular}[c]{@{}c@{}}Feature \\ fusion\end{tabular}    & Resnet-34   & No   & \begin{tabular}[c]{@{}c@{}}RBG\\ Depth\\ NIR\end{tabular}    & \begin{tabular}[c]{@{}c@{}}Feature fusion\\ BCE loss\end{tabular}      \\ \hline
		
		ZKBH    & \begin{tabular}[c]{@{}c@{}}Regression \\ model\end{tabular}  & Resnet-18   & No   & \begin{tabular}[c]{@{}c@{}}RBG\\ Depth\\ NIR\end{tabular}   & \begin{tabular}[c]{@{}c@{}}Data fusion\\ Regression loss\end{tabular}  \\ \hline

		VisionMiracle    & \begin{tabular}[c]{@{}c@{}}Modified \\ shufflenet-V2\\Depth only\\\end{tabular}  & Shufflenet-V2~\cite{ma2018shufflenet}    & No  & Depth   & BCE loss   \\ \hline

		Baseline~\cite{zhang2019dataset,zhang2020casia}    & \begin{tabular}[c]{@{}c@{}} Feature \\ fusion\end{tabular}   & Resnet-18    & No       & \begin{tabular}[c]{@{}c@{}}RBG\\ Depth\\ NIR\end{tabular}   & \begin{tabular}[c]{@{}c@{}}SE fusion\\ BCE loss\end{tabular}   \\ \hline

		GradiantResearch & \begin{tabular}[c]{@{}c@{}}Deep metric \\ learning\end{tabular}    & \begin{tabular}[c]{@{}c@{}}Inception \\ resnet v1~\cite{szegedy2017inception}\end{tabular}    & \begin{tabular}[c]{@{}c@{}}VGGFace2~\cite{cao2018vggface2}\\ GRAD-GPAD~\cite{grad_gpad_gradiant_2019}\end{tabular}     & \begin{tabular}[c]{@{}c@{}}RBG\\ Depth\\ NIR\end{tabular}  & \begin{tabular}[c]{@{}c@{}} Logistic regression\\ensemble \end{tabular}  \\ \hline

		Vipl-bpoic~\cite{wang2019multi}  & \begin{tabular}[c]{@{}c@{}}Attention \\ mechanism~\cite{woo2018cbam}\end{tabular}  & ResNet-18   & No  & \begin{tabular}[c]{@{}c@{}}RBG\\ Depth\\ NIR\end{tabular}   & \begin{tabular}[c]{@{}c@{}}Data fusion\\ Center loss~\cite{wen2016discriminative}\\ BCE loss\end{tabular} \\ \hline

		Massyhnu  & Ensembling    & \begin{tabular}[c]{@{}c@{}}9 softmax \\ classifiers\end{tabular}  & No  & \begin{tabular}[c]{@{}c@{}}RBG\\ Depth\\ NIR\end{tabular}  & \begin{tabular}[c]{@{}c@{}}Colour \\ information fusion\\ BCE loss\end{tabular}   \\ \hline
		
		AI4all  & \begin{tabular}[c]{@{}c@{}}Depth only\end{tabular}     & VGG16~\cite{simonyan2014very}     & No     & Depth    & BCE loss   \\ \hline
		
		Guillaume   & \begin{tabular}[c]{@{}c@{}}Multi-Channel\\ CNN\\No RGB\end{tabular}   & LightCNN~\cite{wu2018light}   & Yes    & \begin{tabular}[c]{@{}c@{}}Depth\\ NIR\end{tabular}      & \begin{tabular}[c]{@{}c@{}}Data fusion\\ BCE loss\end{tabular}  \\ \hline

		Vivi  & \begin{tabular}[c]{@{}c@{}}Dense\\ cross-modal-\\ attention model\end{tabular} & Densenet~\cite{zhu2017densenet}      & Yes       & \begin{tabular}[c]{@{}c@{}}RBG\\ Depth\\ NIR\end{tabular}  & \begin{tabular}[c]{@{}c@{}}Feature fusion\\ Score fusion\\ BCE loss\end{tabular}   \\ \hline
		
	\end{tabular}
	
	\label{table:team-method-summary}
\end{table*}

\textbf{Summary of the participating solutions:}~Table~\ref{table:team-method-summary} summarizes the face PAD solutions of the 13 participating teams and the baseline method. Different from the previous three competitions (\ie, IJCB2011~\cite{chakka2011competition}, ICB2013~\cite{chingovska20132nd} and IJCB2017~\cite{boulkenafet2017competition}), none of the final teams used traditional face PAD methods, such as hand-crafted image quality/texture descriptors~\cite{boulkenafet2016face}, and liveness cues like eye blinking, facial expression changes and mouth movements~\cite{pan2007eyeblink}. Instead, all the submitted face PAD solutions relied on data-driven model-based feature extractors, such as ResNet~\cite{he2016deep} and VGG16~\cite{simonyan2014very}. Furthermore, most of the approaches were multi-modal, combining two or three modalities, while only three teams (Hahaha, VisionMiracle and AI4all) relied on unimodal (depth-based) PAD solution. It can be seen from the last column in Table~\ref{table:team-method-summary} that several kinds of multi-modal fusion strategies (\eg, input-level data fusion, feature-level squeeze and excitation (SE)~\cite{he2016deep} fusion and score-level fusion) were used. In terms of the pre-training data, two teams (VisionLabs and GradiantResearch) leveraged pre-trained models from related face analysis tasks (\eg, face recognition models on CASIA-WebFace~\cite{yi2014learning} and face PAD models on GRAD-GPAD~\cite{grad_gpad_gradiant_2019}) to mitigate the issues with overfitting. It is worth to note that all three top-performing solutions (VisionLabs, ReadSense and Feather) adopted ensemble strategy to aggregate the predictions from multiple variant models.

\begin{table*}[]
	
	\caption{Results and rankings of the stage teams~\cite{liu2019multi} at the final stage. The best results are bolded. ($*$ denotes Vivi that is affiliated with the sponsor and did not participate in the final ranking).}
	
	\centering
	\begin{tabular}{|c|c|c|c|c|c|c|c|c|c|}
		\hline
		\multirow{2}{*}{Team Name} & \multirow{2}{*}{FP} & \multirow{2}{*}{FN} & \multirow{2}{*}{APCER(\%)} & \multirow{2}{*}{NPCER(\%)} & \multirow{2}{*}{ACER(\%)} & \multirow{2}{*}{TPR(\%)@FPR=10e-4}  \\ 
		&                 &                &                &           &         & \\
		\hline
		\hline
		VisionLabs        & \textbf{3}     & 27             & \textbf{0.0074}   & 0.1546             & 0.0810       & \textbf{99.8739}   \\ \cline{1-7}
		ReadSense         & 77             &  \textbf{1}    & 0.1912            & \textbf{0.0057}    & 0.0985        &  99.8052       \\ \cline{1-7}
		Feather           & 48             & 53             & 0.1192            & 0.1392             & 0.1292        & 98.1441             \\ \cline{1-7}
		Hahahaha          & 55             & 214            & 0.1366            & 1.2257             & 0.6812                & 93.1550            \\ \cline{1-7}
		MAC-adv-group     & 825            & 30             & 2.0495            & 0.1718             & 1.1107            & 89.5579             \\ \cline{1-7}
		ZKBH              & 396            & 35             & 0.9838            & 0.2004             & 0.5921               & 87.6618     \\ \cline{1-7}
		VisionMiracle     & 119            & 83             & 0.2956            & 0.4754             & 0.3855           & 87.2094         \\ \cline{1-7}
		GradiantResearch  & 787            & 250            & 1.9551            & 1.4320             & 1.6873            & 63.5493            \\ \cline{1-7}
		Baseline          & 1542           & 177            & 3.8308            & 1.0138             & 2.4223           & 56.8381     \\ \cline{1-7}
		Vipl-bpoic        & 1580           & 985            & 3.9252            & 5.6421             & 4.7836         & 39.5520            \\ \cline{1-7}
		Massyhnu          & 219            & 621            & 0.5440            & 3.5571             & 2.0505           & 29.2990         \\ \cline{1-7}
		AI4all            & 273            & 100            & 0.6782            & 0.5728             & 0.6255            & 25.0601        \\ \cline{1-7}
		Guillaume         & 5252           & 1869           & 13.0477           & 10.7056            & 11.8767             & 0.1595            \\ \cline{1-7}
		Vivi$^{*}$        & 7              & 15             & 0.0173            & 0.0859             & \textbf{0.0516}          & 99.8282             \\ \hline
		
	\end{tabular}
	
	\label{table:team-result-tab}
\end{table*}

\begin{figure}[ht]
	\centering
	\setlength{\textfloatsep}{10pt}
	\includegraphics[scale=0.3]{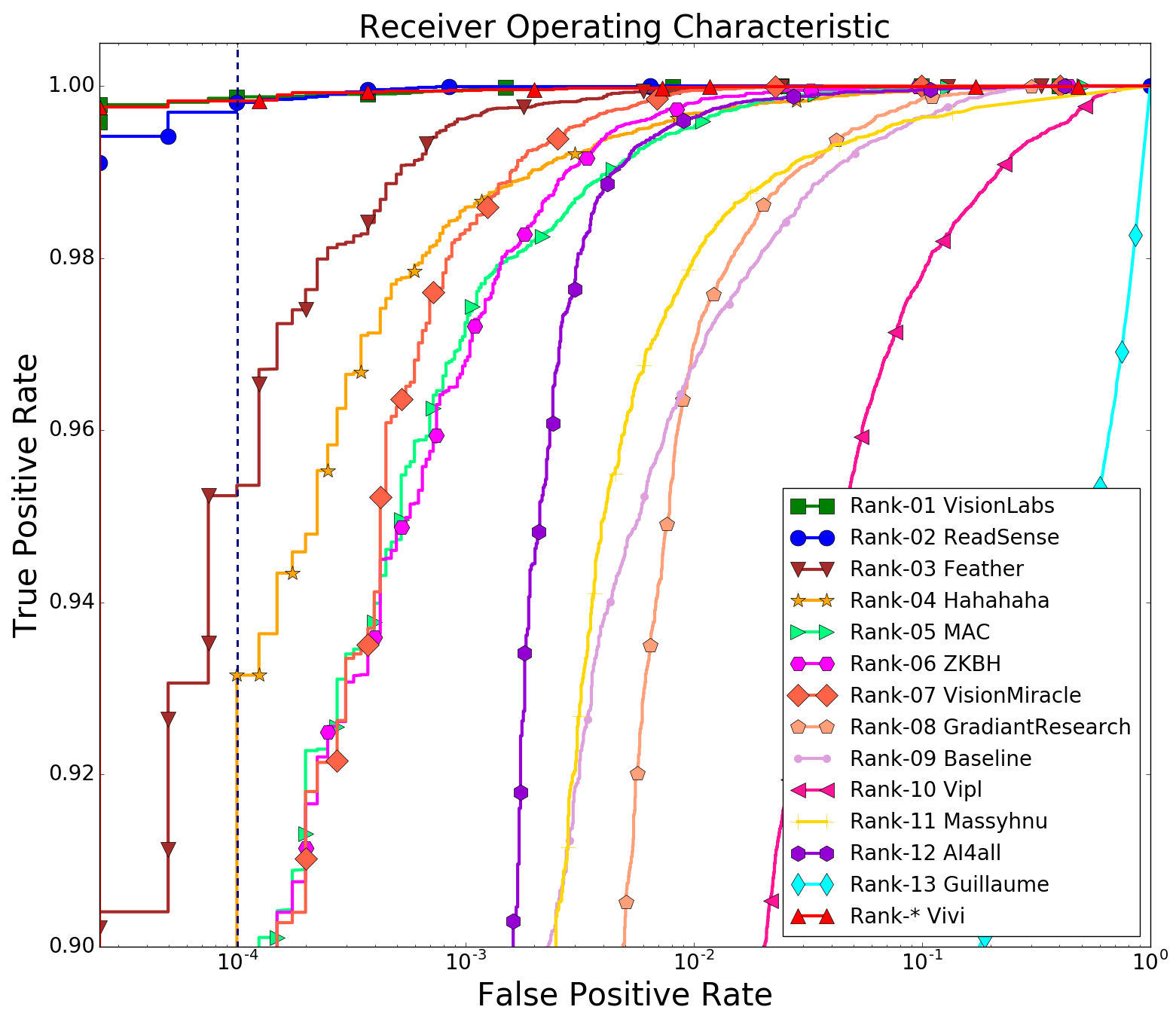}
	\caption{ROC curves of the teams on test set at the final stage~\cite{liu2019multi}.}
	\label{fig:result_pic_valid}
\end{figure}

\textbf{Result analysis:}
The scores and ROC curves of participating teams on the testing partitions are shown in Table~\ref{table:team-result-tab} and Fig.~\ref{fig:result_pic_valid}, respectively. It can be observed that the winning team (VisionLabs) achieved TPR$99.8739\%$@FPR=$10^{-4}$, and the FN = $27$ and FP = $3$ on the test set. In fact, different application scenarios have different requirements for each indicator. For example, to meet the higher security needs of an access control system, the FP is required to be as small as possible. With respect to this criterion, the VisionLabs performed very well as only three attack samples were misclassfied as the bona fide. In contrast, a small FN value is more important in the case of finding suspects where the team ReadSense achieved best result (FN=1) due to the effectiveness of local patch inputs. In overall, the first eight teams were performing better than the baseline method~\cite{zhang2019dataset,zhang2020casia} in terms of FP and TPR@FPR=$10^{-4}$, indicating the valuable outputs and insightful solutions in this challenge.

As shown in Table~\ref{table:team-result-tab}, the results of the three top-performing teams on the test set were clearly superior compared with the other teams. Combining Table~\ref{table:team-method-summary} with Table~\ref{table:team-result-tab}, we can conclude that ensemble learning performed more robustly compared to single-model based solutions under the same conditions. The ROC curves visualization of all participating teams submissions are illustrated in Fig.~\ref{fig:result_pic_valid}. It can be seen that three teams (\ie, VisionLabs, ReadSense, and Vivi) were significantly better than other teams on test set. For instance, the TPR@FPR=$10^{-4}$ values of these three teams are relatively close to each other and superior compared to the other teams. In consideration of the characteristics of modalities (\eg, the colour data is rich in details; the depth data is sensitive to the distance, and the NIR data measures better face-specific reflectance), the teams Vivi and Vipl-bpoic introduced the attention mechanisms into face PAD task, which enforced the models to focus on more informative regions in the different modalities. Similarly, the team Feather used a cascaded architecture with two sub-networks to study the CASIA-SURF dataset with two modalities, when depth and NIR data were examined subsequently by each network. Some teams considered also face landmarks (\textit{e.g.}, Hahahaha) and color space conversions (\textit{e.g.}, MAC-adv-group, Massyhnu) for PAD. Instead of conventional binary classification based PAD model, the team ZKBH constructed a regression model to supervise the model to learn local cues in the eye regions. In order to generalize better to unseen attacks, the team GradiantResearch reformulated face PAD as an anomaly detection problem using deep metric learning.

\textbf{Discussion:}~Although most of the proposed solutions achieved superior performance compared with the provided multi-modal SE fusion based baseline, there were still some limitations in the CVPR2019 challenge. Originally, the main research question was to explore new efficient multi-modal fusion schemes for combining colour, depth and NIR modalities. However, no novel or otherwise insightful multi-modal fusion strategies were proposed in the end. Most of the teams applied very simple data-level and score-level fusion with greedy searching manner, which is likely to fail when evaluating a method on an unknown multi-modal dataset. Furthermore, the two top-performing teams adopted the same feature-level SE fusion strategy as the baseline method, which again was not very fruitful from the multi-modal challenge point of view. Many of the top-performing solutions exploited ensemble of multiple models to boost the performance. However, while pushing the efficiency, ensembling also increases the complexity of the whole solution, which is not practical in real-world conditions, especially considering the limitations with mobile and embedded platforms. Finally, some solutions were considerably inspired by the human-observed priors in the CASIA-SURF dataset (\eg, the apparent discrepancy in the eye and nose regions between bona fide and attack samples), which were easily fooled by the cut paper attacks with similar shapes in these regions. Based on the aforementioned observations, the problem of designing more generalized multi-modal face PAD solutions capturing specific intrinsic fidelity characteristics between bona fide and attack samples remains an open issue.

\subsection{Cross-ethnicity Face Anti-spoofing Recognition Challenge (CVPR2020)}
\label{sec:com2}


The racial bias in face PAD methods was not explicitly explored until it was demonstrated in~\cite{liu2021casia} that the PAD performance of deep models can vary widely on the test samples with unseen ethnicity. To alleviate the racial bias and ensure the reliability of face PAD methods among different populations, the CASIA-SURF Cross-ethnicity Face Anti-Spoofing (CeFA) dataset~\cite{liu2021casia} along with the Chalearn Cross-ethnicity Face Anti-Spoofing Recognition Challenge~\cite{cvpr2020challenge} were established. 

The cross-ethnicity face PAD challenge comprised unimodal (\ie, colour) and multi-modal (\ie, colour, depth and NIR) competition tracks, which were collocated with the Workshop on Media Forensics\footnote{\url{https://sites.google.com/view/wmediaforensics2020/home}} at CVPR2020. Similarly to the previous multi-modal challenge, both the unimodal\footnote{\url{https://competitions.codalab.org/competitions/22151}} and multi-modal\footnote{\url{https://competitions.codalab.org/competitions/22036}} tracks were run simultaneously using the CodaLab platform. The competition attracted $340$ teams in the development stage, with $11$ and eight teams finally entering the actual evaluation stage for the unimodal and multi-modal face PAD tracks, respectively. Summary of the names and affiliations of teams that entered the final stage as well as their final rankings are shown in Tables~\ref{affiliations_Single} and~\ref{affiliations_multi} for the unimodal and multi-modal tracks, respectively. From the tables, it can be seen that most participants came from  industrial institutions, indicating the increasing need for reliable and robust PAD systems for practical AFR applications. Interestingly, the team VisionLabs was not only the winner of the unimodal track of the CVPR2020 challenge, but also the winner of the earlier multi-modal CVPR2019 challenge~\cite{ChallengeCVPR2019}. In addition, the team BOBO from University of Oulu (the authors' team) proposed novel central difference convolution (CDC)~\cite{yu2020searching,yu2020fas} and contrastive depth loss (CDL)~\cite{wang2020deep} methods for feature learning, achieving the first and second place in multi-modal and unimodal tracks, respectively.

\begin{table}[!b]
\caption{Names, affiliations and rankings of the participating systems in the unimodal track~\cite{cvpr2020challenge}.\label{affiliations_Single}}
{
\centering
\begin{tabular}{|c|c|c|}
		\hline
		Ranking & Team Name        & Affiliation                                  \\ \hline
		\hline
		1 & VisionLabs              & VisionLabs                                 \\ \hline
		2 & BOBO                    & Zitong Yu, University of Oulu                                       \\ \hline
		3 & Harvest                 & Jiachen Xue, Horizon                                        \\ \hline
		4 & ZhangTT                 & Zhang Tengteng, CMB                                         \\ \hline
		5 & Newland\_tianyan        & Xinying Wang, Newland Inc.                              \\ \hline
		6 & Dopamine                & Wenwei Zhang, huya                                     \\ \hline
		7 & IecLab                  & Jin Yang, HUST                                \\ \hline
		8 & Chuanghwa Telecom Lab.  & Li-Ren Hou, Chunghwa Telecom                                                             \\ \hline
		9 &  Wgqtmac                 & Guoqing Wang, ICT                                                       \\ \hline
		10 &  Hulking                & Yang, Qing, Intel                                                          \\ \hline
		11 &  Dqiu                   & Qiudi                                        \\ \hline
	\end{tabular}

}{}
\end{table}

\begin{table}[!b]
	\caption{Names, affiliations and rankings of the participating systems in the multi-modal track~\cite{cvpr2020challenge}.\label{affiliations_multi}} {
	\centering
	\begin{tabular}{|c|c|c|}
		\hline
		Ranking & Team Name        & Affiliation                                  \\ \hline
		\hline
		1 & BOBO  & Zitong Yu, University of Oulu                                      \\ \hline
		2 & Super  & Zhihua Huang, USTC                                      \\ \hline
		3 & Hulking &  Qing Yang, Intel                                          \\ \hline
		4 & Newland\_tianyan  & Zebin Huang, Newland Inc.                                          \\ \hline
		5 & ZhangTT  &Tengteng Zhang, CMB                           \\ \hline
		6 & Harvest  &Yuxi Feng, Horizon                                     \\ \hline
		7 & Qyxqyx  &Yunxiao Qin, NWPU                                      \\ \hline
		8 & Skjack  &Sun Ke, XMU                                                                    \\ \hline
	\end{tabular}
	
}{}
\end{table}

\subsubsection{Dataset}

\begin{figure}[t]
	\begin{center}
	\includegraphics[width=1.0\linewidth]{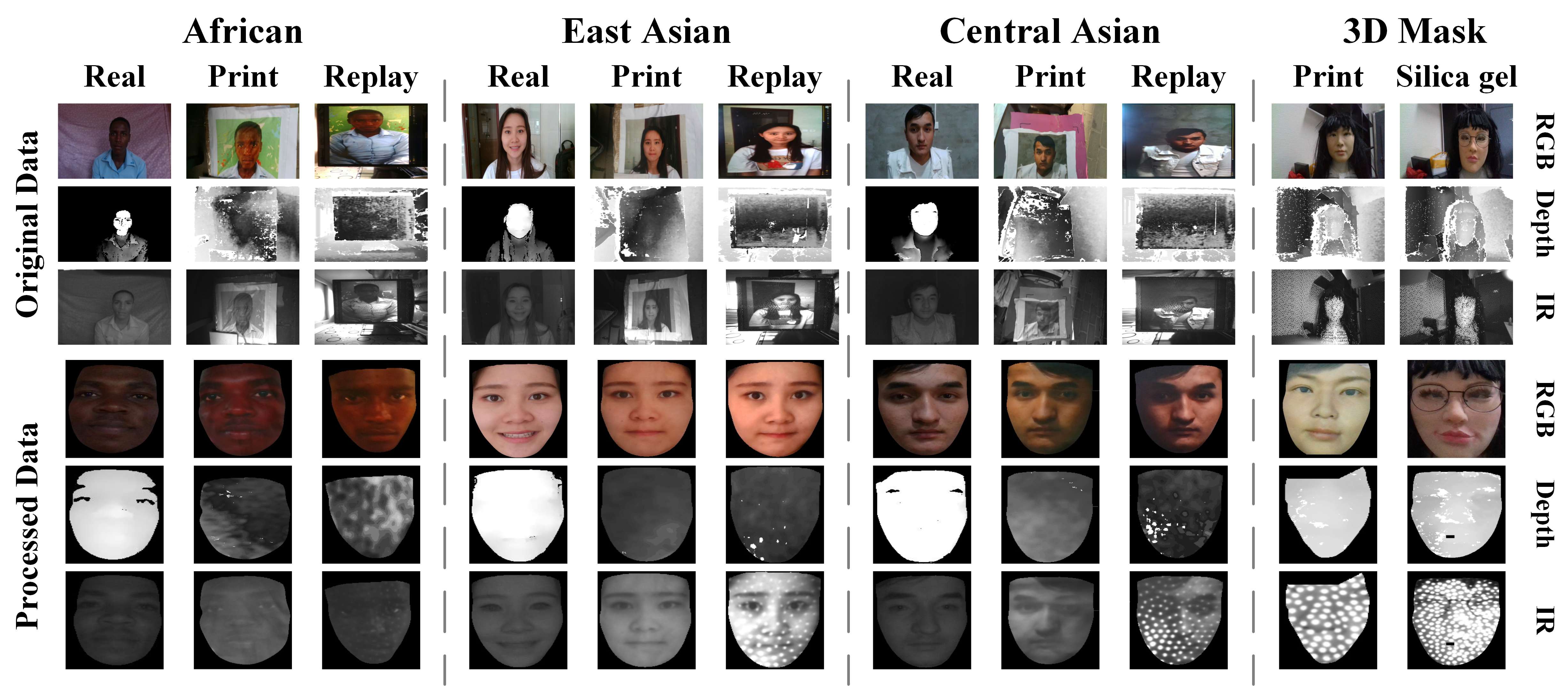}
	\end{center}
	\caption{Samples from the CASIA-SURF CeFA dataset~\cite{liu2021casia}, consisting of $1,607$ subjects, three different ethnicities (\ie, Africa, East Asia, and Central Asia), four PA types (\ie, print, video-replay, 3D print and silica gel mask) and three modalities (\ie, colour, depth, and NIR).}
	\label{samp_data_CeFA}
\end{figure}

The CASIA-SURF CeFA~\cite{liu2021casia} was the largest face anti-spoofing dataset at the time of the CVPR2020 competition, covering three ethnicities (\ie, Africa, East Asia and Central Asia), three modalities (\ie, colour, depth and NIR), $1,607$ subjects, and four different of PA types (\ie, prints, video-replays, 3D print and silica gel masks). The multi-modal videos were captured using Intel RealSense SR300 camera with resolution of 1280 × 720 pixels for each video frame at 30 frames per second. The data was pre-processed in similar way as in the previous CVPR2019 multi-modal challenge~\cite{liu2019multi} (see, Section \ref{CASIA-SURF_dataset}). The CASIA-SURF CeFA was the first public dataset designed for exploring also the racial bias of face PAD methods. Some samples of the CASIA-SURF CeFA dataset are shown in Fig.~\ref{samp_data_CeFA}. 

The main motivation of CASIA-SURF CeFA dataset is to serve as a benchmark allowing evaluation of the generalization of PAD methods across different ethnicities, PAIs and modalities under varying scenarios using four specific protocols: 

$\bullet$ \textbf{Protocol 1:} Cross-ethnicity generalization of PAD methods is evaluated by using one ethnicity for training and validation, while the two remaining ones are used as unseen ethnicities for testing. 

$\bullet$ \textbf{Protocol 2:} Cross-PAI generalization of PAD methods is evaluated by using print or video-replay attack for training and validation, while the remaining three attacks are used as unknown PA types for testing. 

$\bullet$ \textbf{Protocol 3:} Cross-modality generalization of PAD methods is evaluated by using one modality for training and validation, while the two remaining ones are used as unknown modalities for testing. 

$\bullet$ \textbf{Protocol 4:} Cross-ethnicity and cross-PAI generalization of PAD methods is evaluated simultaneously by combining the first two protocols, \ie, using one ethnicity and PAI for training and validation, while the remaining two ethnicities as well as three PAIs are used for testing.

\begin{table}[]
\caption{Protocols and statistics of the Protocol $4$ of the CASIA-SURF CeFA~\cite{liu2021casia} dataset, where 'A', 'C', and 'E' denote Africa, Central Asia, and East Asia, respectively.\label{tab:PandS}}

\centering
{
\scalebox{1.0}{
\begin{tabular}{@{}ccccccccccc@{}}
\toprule
\multirow{2}{*}{Subset} &\multirow{2}{*}{Subjects} & \multicolumn{3}{c}{Ethnicity} &\multirow{2}{*}{PAIs} &\multicolumn{3}{c}{\# images (RGB)} \\
\cmidrule(lr){3-5}
\cmidrule(l){7-9}
& (one ethnicity) & 4\_1 & 4\_2  & 4\_3
& & 4\_1 & 4\_2  & 4\_3
\\ \midrule
 Train
& 1-200  & A   & C  & E  & Video-Replay
& 33,713  & 34,367      & 33,152     \\
Valid & 201-300  & A   & C   & E  & Video-Replay  & 17,008 & 17,693 & 17,109     \\
Test  & 301-500  & C\&E & A\&E & A\&C & Print+Mask  & 105,457  & 102,207 & 103,420 \\
\bottomrule
\end{tabular}
}
}{}
\end{table}

The most challenging Protocol $4$ was adopted for ranking the methods in both unimodal and multi-modal tracks of the competition. As shown in Table~\ref{tab:PandS}, this protocol consists of three subsets: training, validation, and test sets, containing $200$, $100$, and $200$ subjects for each ethnicity, respectively. Note that the remaining 107 subjects correspond to the 3D masks attacks. Since there are three ethnicities in CASIA-SURF CeFA, in total three sub-protocols (\ie, $4\_1$, $4\_2$ and $4\_3$ in Table~\ref{tab:PandS}) were adopted in the CVPR2020 challenge. In addition to the racial variation, the unknown PAIs introduced in the test sets made the competition even more challenging.

\subsubsection{Evaluation Protocol and Metrics}

The challenge comprised development and final stages. The detailed protocols are described as follows.

\textbf{Protocol in development phase:}~(\emph{Dec. 13, 2019 - March 1, 2020}). During the development phase, the participants had access to the labeled training set and unlabeled validation set. Since the Protocol 4 of the CASIA-SURF CeFA dataset used in this competition comprised three sub-protocols (see Table~\ref{tab:PandS}), the participants first needed to train a model for each sub-protocol and then predict the scores for each corresponding validation set. Finally, the participants had to merge the predicted scores of the three sub-protocols and submit the resulting final scores to the CodaLab platform, where an immediate response was seen in the public leaderboard.

\textbf{Protocol in final phase:}~(\emph{March 1, 2020 - March 10, 2020}). During the final phase, the labeled validation set and the unlabeled testing set were released. The participants could first utilize the labels of the validation set for model selection to improve the generalization on the test data. All results of the three sub-protocols were made publicly available online in terms of APCER, BPCER, and ACER. Like with the OULU-NPU dataset~\cite{Boulkenafet2017OULU} used in the IJCB2017 competition \cite{boulkenafet2017competition}, the mean and variance of evaluated metrics across the three different sub-protocols were calculated and included in the final results.

Note that in order to fairly compare the performance of different submitted systems, the CVPR2020 challenge did not allow the use of external training datasets or pre-trained models. All participants were encouraged to release their source codes under feasible licenses and to provide a fact sheet describing their solution. All codes would be re-run and verified by the organizing team after the final submission phase. For the sake of reproducibility and fairness, the final ranking of the teams was based on the verified results.

\textbf{Evaluation metrics:}~Similarly to the previous CVPR2019 competition \cite{ChallengeCVPR2019}, also in the CVPR2020 challenge, the standardized ISO/IEC 30107-3~\cite{ACER} metrics (\ie, APCER, BPCER and ACER) were considered as the main evaluation metrics (see, Section \ref{sec:metrics}). Also, ROC was included for visualization purposes and additional result analysis. The final rankings were based on the ACER metric on the test set because it has been widely used for evaluating the performance of face PAD systems in the literature and majority of the previous face PAD competitions. The ACER threshold was determined by calculating the Equal Error Rate (EER) operating point on the validation set.

\subsubsection{Results and Discussion}

In this section, we first summarize the methods and report the results of the unimodal track, and then analyze the solutions as well as the results of the multi-modal track. Finally, we provide general discussion on the proposed algorithms and competition.

\textbf{Solutions of the unimodal (colour) track:}~Table~\ref{tab:single_solutions} summarizes the face PAD solutions of the teams participated in the unimodal track. The source codes of ten teams, including VisionLabs\footnote{\url{https://github.com/AlexanderParkin/CASIA-SURF_CeFA}}, BOBO\footnote{\url{https://github.com/ZitongYu/CDCN/tree/master/FAS_challenge_CVPRW2020}}, Harvest\footnote{\url{https://github.com/yueyechen/cvpr20}}, ZhangTT\footnote{\url{https://github.com/ZhangTT-race/CVPR2020-SingleModal}}, Newland-tianyan\footnote{\url{https://github.com/XinyingWang55/RGB-Face-antispoofing-Recognition}}, Dopamine\footnote{\url{https://github.com/xinedison/huya_face}}, IecLab\footnote{\url{https://github.com/1relia/CVPR2020-FaceAntiSpoofing}}, Chungwa-Telecom\footnote{\url{https://drive.google.com/open?id=1ouL1X69KlQEUl72iKHl0-_UvztlW8f_l}}, Wgqtmac\footnote{\url{https://github.com/wgqtmac/cvprw2020.git}}, and Hulking\footnote{\url{https://github.com/muyiguangda/cvprw-face-project}} were made publicly available. It was not surprising that every team adopted end-to-end learning based approaches due to the strong representation capacity of modern deep models. Regarding the model inputs, most of the teams used the provided facial colour images directly, while the winning team VisionLabs considered two kinds of pre-processing methods for dynamic inputs (\ie, optical flow~\cite{horn1981determining} and rank pooling~\cite{fernando2016rank} images). As for the backbone networks, only the team Dopamine adopted spatio-temporal 3D convolutional neural network (CNN) model, while the others relied on 2D CNNs (mostly ResNet). Most of the solutions treated face PAD as a binary classification problem via simple BCE loss, but few teams (\ie, BOBO, Harvest, and ZhangTT) considered pixel-wise depth loss, temporal continuous L1 regression loss and multi-class softmax CE loss, respectively. It is interesting to see from the last two columns of Table \ref{tab:single_solutions} that most of the solutions leveraged dynamic cues, while did not adopt complex model ensemble strategy. 

\begin{table*}[]
\caption{Summary of the top-performing solutions in the unimodal track of the CVPR2020 challenge, where 'S/D' indicates Static/Dynamic, 'OF' and 'RP' for optical flow~\cite{horn1981determining} and rank pooling~\cite{fernando2016rank}, 'BCE', 'CDC', and 'CDL' binary cross-entropy, central difference convolution, and contrastive depth loss, respectively. \label{tab:single_solutions}}
{
\scalebox{0.8}{
\begin{tabular}{@{}clcccccccc@{}}
\toprule
Team Name & Method (keywords)  & Input & Backbone & Loss function & S/D & Ensemble \\ \midrule

VisionLabs
&  Creating artificial modalities & OF+RP & SimpleNet~\cite{parkin2020creating}    & BCE loss  &  D & No \\ \midrule

BOBO
& CDC, CDL, Attention
& RGB & CDCN~\cite{yu2020searching}   &  Depth loss & S  & Yes \\ \midrule

Harvest
& Motion-aware labels & RGB   & ResNet101 & L1 loss  & D & Yes \\ \midrule

ZhangTT
& Quality tensor
& Grayscale  & ResNet~\cite{he2016deep}  & 4-class CE loss & D & No \\ \midrule

Newland-tianyan
& Subtracted neighborhood mean
& RGB  & 5-layer network  & BCE loss & D  & No \\ \midrule

Dopamine
& Multi-task learning
& RGB & ResNet100  & BCE + face ID loss & S & No\\ \midrule

IecLab
&  Authenticity+expression features
& RGB &  3DResNet~\cite{hara3dcnns} & BCE loss & D & No \\ \midrule

Chunghwa-Telecom
&  Bag of local features
& RGB & MIMAMO-Net~\cite{deng2020mimamo}   & BCE loss & S  & Yes\\ \midrule

Wgqtmac
& Warmup strategy
& RGB & ResNet18 & BCE loss & S & No \\ \midrule

Hulking
& Frame vote module 
& RGB patch & PipeNet~\cite{fernando2016rank}  & BCE loss  & D & No\\ \midrule

Dqiu
& -
& RGB & ResNet50 &  BCE loss & S & No\\ \midrule

Baseline
& Hybrid feature fusion
& RGB+RP & SD-Net~\cite{liu2021casia} & BCE loss & D & No \\ \bottomrule
\end{tabular}
}
}{}
\end{table*}

\textbf{Results of the unimodal (colour) track:}~
The final results of the $11$ participated teams are shown in Table~\ref{tab:single_results}. The final ranking was based on the mean ACER computed over the three sub-protocols. The EER thresholds from validation set are also reported in Table~\ref{tab:single_results}. The threshold values for the best-performing algorithms were either extremely large (\eg, more than 0.9 for BOBO) or small (\eg, 0.01 for Harvest). In contrast, VisionLabs's algorithm was more stable with threshold values around 0.5 and achieving the highest accuracy in detecting the PA samples (APCER = $2.72\%$), while Wgqtmac's algorithm obtained the best results in terms of BPCER ($0.66\%$). In overall, the first ten teams were performing better than the baseline method~\cite{liu2021casia} in terms of ACER. The top three teams obtained excellent ACER values below $10\%$, and the team VisionLabs achieved the first place with a clear margin.

\begin{figure*}[t]
	\begin{center}
	\includegraphics[width=1.0\linewidth]{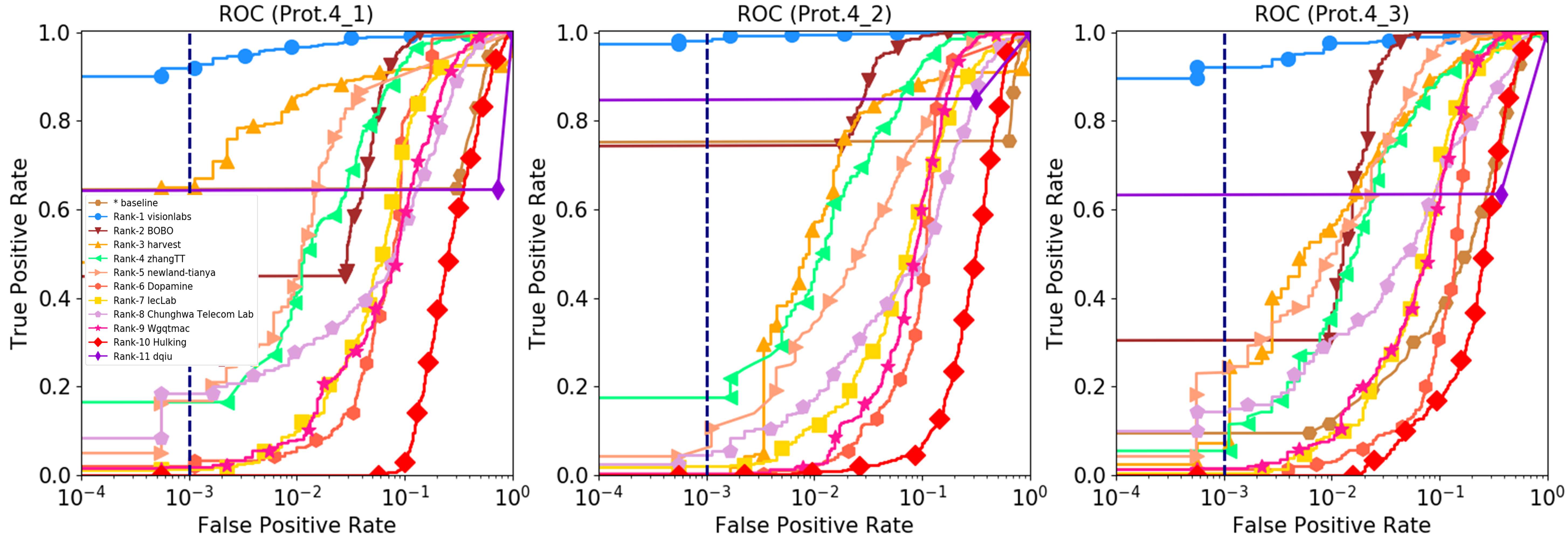}
	\end{center}
	\caption{The ROC curves for the $12$ teams participating the unimodal track of the CVPR2020 challenge~\cite{liu2021casia}. From left to right are the ROC curves for protocol 4\_1, 4\_2 and 4\_3, respectively.}
	\label{fig:sroc}
\end{figure*}

The ROC curves of the three sub-protocols are given in Fig.~\ref{fig:sroc} to further analyze the trade-off between FPR and TPR, \ie, tuning the operating point according to the requirements of a given real-world application. The results of the winning team VisionLabs (blue curve) on all three sub-protocols are clearly superior compared to others, indicating the advantages of optical flow based motion clues and rank pooling images in improving the generalization performance. However, the TPR value of the remaining teams decreases rapidly as the FPR reduces (\textit{e.g.}, TPR@FPR=$10^{-3}$ values for these teams are almost zero). In addition, although the ACER of the team Harvest was worse than that of the team BOBO, its TPR@FPR=$10^{-3}$ was significantly better than that of BOBO. It was mainly because the values of FP and FN samples for the team Harvest were relatively close to each other (see Table~\ref{tab:single_results}).

\begin{table*}[t]
\centering
\caption{The results of the unimodal track of the CVPR2020 challenge \cite{liu2021casia}. Avg$\pm$Std denotes the mean and variance computed across the three sub-protocols. The best results are shown in bold.\label{tab:single_results}}
{
\scalebox{0.9}{
\begin{tabular}{@{}clcccccccc@{}}
\toprule
Team Name  & Threshold & FP  & FN & APCER(\%) & BPCER(\%) & ACER(\%) & Rank \\ \midrule

VisionLabs
&   0.34$\pm$0.48 & 2$\pm$2      & 21$\pm$9  & \textbf{0.11$\pm$0.11} & 5.33$\pm$2.37   & \textbf{2.72$\pm$1.21} & 1 \\ \midrule

BOBO

& 0.97$\pm$0.02 & 129$\pm$67   & 10$\pm$2    & 7.18$\pm$3.74  & 2.50$\pm$0.50  & 4.84$\pm$1.79  &  2\\ \midrule

Harvest
& 0.01$\pm$0.00 & 85$\pm$47    & 55$\pm$10  & 4.74$\pm$2.62 & 13.83$\pm$2.55  & 9.28$\pm$2.28  & 3 \\ \midrule

ZhangTT

& 0.9  & 97$\pm$37  & 75$\pm$31 & 5.40$\pm$2.10 & 18.91$\pm$7.88  & 12.16$\pm$2.89  &   4\\ \midrule

Newland-tianyan

& 0.67$\pm$0.11 & 282$\pm$239  & 44$\pm$62   & 15.66$\pm$13.33& 11.16$\pm$15.67   & 13.41$\pm$3.77  &  5\\ \midrule

Dopamine

 & 0.07$\pm$0.11 & 442$\pm$168  & 10$\pm$12   & 24.59$\pm$9.37 & 2.50$\pm$3.12 & 13.54$\pm$3.95 & 6\\ \midrule

IecLab

& 0.40$\pm$0.07 & 597$\pm$103  & 24$\pm$2  & 33.16$\pm$5.76   & 6.08$\pm$0.72  & 19.62$\pm$2.59  &  7\\ \midrule

Chunghwa-Telecom

& 0.86$\pm$0.06 & 444$\pm$93   & 76$\pm$34  & 24.66$\pm$5.16                          & 19.00$\pm$8.69  & 21.83$\pm$1.82 & 8\\ \midrule

Wgqtmac

& 0.80$\pm$0.22 & 928$\pm$310  & 2$\pm$3  & 51.57$\pm$17.24   & \textbf{0.66$\pm$0.94} & 26.12$\pm$8.15 & 9 \\ \midrule

Hulking
 & 0.76$\pm$0.08 & 810$\pm$199  & 78$\pm$53  & 45.00$\pm$11.07 & 19.50$\pm$13.27  & 32.25$\pm$3.18  & 10\\ \midrule

Dqiu
& 1.00$\pm$0.00 & 849$\pm$407  & 116$\pm$48 & 47.16$\pm$22.62
& 29.00$\pm$12.13 & 38.08$\pm$15.57 & 11\\ \midrule

Baseline

& 1.00$\pm$0.00 & 1182$\pm$300 & 30$\pm$25  & 65.66$\pm$16.70     & 7.58$\pm$6.29 & 36.62$\pm$5.76 &  \\ \bottomrule
\end{tabular}
}
}{}
\end{table*}

\textbf{Solutions of the multi-modal track:}~Table~\ref{tab:multi_solutions} summarizes the face PAD solutions of the teams participated in the multi-modal track. The source codes of seven teams, including BOBO, Super\footnote{\url{https://github.com/hzh8311/challenge2020_face_anti_spoofing}}, Hulking\footnote{\url{https://github.com/ZhangTT-race/CVPR2020-SingleModal}}, Newland-tianyan\footnote{\url{https://github.com/Huangzebin99/CVPR-2020}}, ZhangTT\footnote{\url{https://github.com/ZhangTT-race/CVPR2020-MultiModal}}, Qyxqyx\footnote{\url{https://github.com/qyxqyx/FAS_Chalearn_challenge}}, and Skjack\footnote{\url{https://github.com/skJack/challange.git}} were made publicly available. Most of the teams exploited all the three modalities (colour, depth and NIR) for feature and score level fusion, except for the teams ZhangTT and Harvest who considered only depth and NIR modalities. There were no teams using data level fusion strategy. As for the architectures and loss functions, teams BOBO and Qyxqyx adopted MM-CDCN~\cite{yu2020multi} and DepthNet~\cite{Liu2018Learning} with pixel-wise supervision, while the other teams relied on ResNet with BCE loss. In contrast to the unimodal track, the use of static cues and model ensembling were popular in the multi-modal track.

\begin{table*}[]
\caption{Summary of the top-ranked solutions of the multi-modal track in the CVPR2020 challenge. \label{tab:multi_solutions}}
{
\scalebox{0.8}{
\begin{tabular}{clcccccccc}
\toprule
Team Name   &  Modality   & Fusion  & Backbone    & Loss function &  S/D    & Ensemble    \\ \midrule

BOBO
& RGB, Depth, NIR
& Feature \& score level  & MM-CDCN~\cite{yu2020multi} & Depth loss   & S & Yes        \\ \midrule

Super
& RGB, Depth, NIR
 &  SE fusion in feature level & ResNet34/50 & BCE loss & S & Yes     \\ \midrule

Hulking
& RGB, Depth, NIR
& Feature level& PipeNet~\cite{yang2020pipenet}   & BCE loss  & D    & No   \\ \midrule

Newland-tianyan
& Grayscale, Depth, NIR
& Score level & Resnet9      &BCE loss & S & No     \\ \midrule

ZhangTT
&  Depth, NIR
& Feature level   &      ID-Net     & BCE loss       & S      &   Yes       \\ \midrule

Harvest
& NIR
& No fusion   &   -        & Triplet loss         & S    &      No      \\ \midrule

Qyxqyx
& RGB, Depth, NIR
& Score level & DepthNet~\cite{Liu2018Learning}     & BCE+BinaryMap loss          & S      & Yes        \\ \midrule

Skjack
& RGB, Depth, NIR
& Feature level &  Resnet9 & BCE loss   & S    &    No           \\ \midrule

Baseline
& RGB, Depth, NIR, RP
& Feature level   & PSMM-Net~\cite{liu2021casia}   & BCE loss   & D & No                 \\ \bottomrule
\end{tabular}
}
}{}
\end{table*}

\begin{table*}[]
\centering
\caption{The results of the multi-modal track in the CVPR2020 challenge~\cite{liu2021casia}. Avg$\pm$Std indicates the mean and variance across the three folds and the best results are shown in bold.\label{tab:multi_results}}
{
\scalebox{0.9}{
\begin{tabular}{clcccccccc}
\toprule
Team Name      & Threshold         & FP             & FN    & APCER(\%)  & BPCER(\%)     & ACER(\%)      & Rank      \\ \midrule

BOBO
& 0.95$\pm$0.02 & 19$\pm$11      & 4$\pm$2   & 1.05$\pm$0.62          & \textbf{1.00$\pm$0.66} & \textbf{1.02$\pm$0.59} &           1         \\ \midrule

Super
 & 1.0$\pm$0.00  & 11.33$\pm$7.76 & 11$\pm$6  & 0.62$\pm$0.43          & 2.75$\pm$1.50          & 1.68$\pm$0.54          &        2            \\ \midrule

Hulking
& 0.98$\pm$0.02 & 58$\pm$35      & 4$\pm$4   & 3.25$\pm$1.98          & 1.16$\pm$1.12          & 2.21$\pm$1.26          &        3            \\ \midrule

Newland-tianyan
& 1.00$\pm$0.00 & 4$\pm$4        & 17$\pm$12 & \textbf{0.24$\pm$0.25} & 4.33$\pm$3.12          & 2.28$\pm$1.66          &         4           \\ \midrule

ZhangTT
& 0.87$\pm$0.07 & 56$\pm$51      & 17$\pm$17 & 3.11$\pm$2.87          & 4.41$\pm$4.25          & 3.76$\pm$2.02          &          5          \\ \midrule

Harvest
& 0.92$\pm$0.04 & 104$\pm$84     & 13$\pm$12 & 5.77$\pm$4.69          & 3.33$\pm$3.21          & 4.55$\pm$3.82          &          6          \\ \midrule

Qyxqyx
& 0.95$\pm$0.05 & 92$\pm$142     & 26$\pm$23 & 5.12$\pm$7.93          & 6.66$\pm$5.86          & 5.89$\pm$4.04          &          7          \\ \midrule

Skjack
& 0.00$\pm$0.00 & 1012$\pm$447   & 47$\pm$45 & 56.24$\pm$24.85        & 11.75$\pm$11.37        & 33.99$\pm$7.08         &         8           \\ \midrule

Baseline
& 0.39$\pm$0.52 & 872$\pm$463    & 62$\pm$43 & 48.46$\pm$25.75        & 15.58$\pm$10.86        & 32.02$\pm$7.56         &                    \\ \bottomrule
\end{tabular}
}
}{}
\end{table*}


\textbf{Results of the multi-modal track:}~
The results of the eight teams participating in the final stage are shown in Table~\ref{tab:multi_results}. The team BOBO team achieved the best performance in terms of $BPCER=1.00\%$ and $ACER=1.02\%$, and the team Super ranked second with a minor margin $ACER=1.68\%$. It is worth noting that the team Newland-tianyan achieved the best results in terms of APCER ($0.24\%$). Similarly to the unimodal track, most of the participating teams had relatively large EER thresholds calculated on the validation set, especially the teams Super and Newland-tianyan with threshold values of $1.0$, indicating that the samples would be easily classified as anomalies. In addition, it can be seen that the ACER values of the four top teams were $1.02\%$, $1.68\%$, $2.21\%$ and $2.28\%$, all outperforming the best performance ($2.72\%$ ) reported in the unimodal track. This suggests that the additional modalities are indeed useful in improving robustness of face PAD under the challenging cross-ethnicity and cross-PAI conditions.

\begin{figure*}[t]
	\begin{center}
	\includegraphics[width=1.0\linewidth]{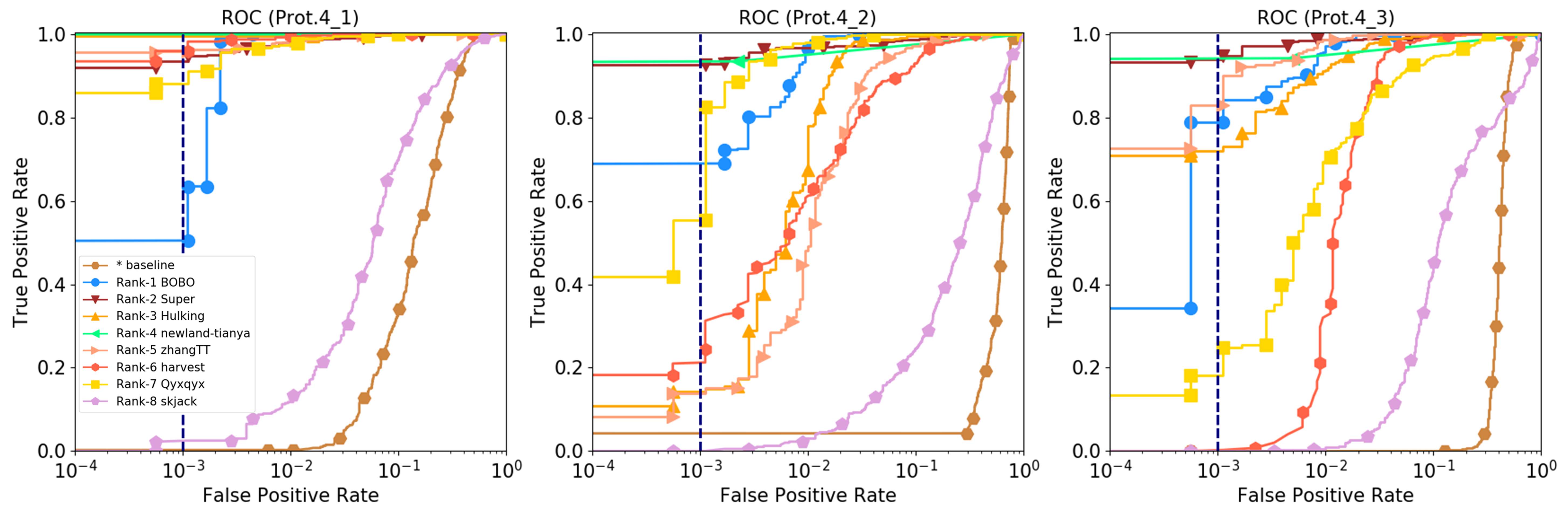}
	\end{center}
	\caption{The ROC curves of nine teams in the multi-modal track~\cite{liu2021casia}. From left to right are the ROCs on protocol 4\_1, 4\_2 and 4\_3, respectively.}
	\label{fig:mroc}
\end{figure*}

The ROC curves of the solutions in multi-modal track are shown in Fig.~\ref{fig:mroc}. From the Table~\ref{tab:multi_results} and Fig.~\ref{fig:mroc}, we can find that even though the ACER values of the top two algorithms were relatively close, the stability of the team Super (brown curve) is better than the BOBO (blue curve). For instance the TPR@FPR=$10^{-3}$ values for Super and Newland-tianyan were better than that of BOBO on all three sub-protocols. In other words, compared with the BCE loss based solution, the depth-wise supervision in team BOBO's solutions might cause larger bias between metrics ACER and TPR@FPR=$10^{-3}$.

\textbf{Discussion:}~From Tables~\ref{tab:single_results} and~\ref{tab:multi_results} of the competition results, we can find that the EER thresholds computed on the validation set for both unimodal and the multi-modal track were generally high, indicating that the proposed algorithms might easily make over-confident or biased decisions. The reason behind this might be two-fold: 1) the biased distributions of the CASIA-SURF CeFA dataset, \eg, as the environments for the attack samples were more diverse, while bona fide samples were usually recorded indoor, and 2) the lack of generalization when the algorithm faces unknown PA types and ethnicities. Moreover, rethinking the evaluation metric for ranking the solution is necessary. It can be seen from the both two tracks that some solutions (\eg, the team BOBO) could achieve excellent ACERs but, on the other hand, unsatisfying ROC curves, especially at operating points with low FPR.

\subsection{CelebA-Spoof Challenge on Face Anti-Spoofing (ECCV2020)}
\label{sec:com3}


Despite both the CVPR2019~\cite{ChallengeCVPR2019} and CVPR2020~\cite{cvpr2020challenge} challenges were successful in benchmarking the generalization of unimodal and multi-modal face PAD methods in challenging settings, the amount of data (number of images $\textless 150,000$ and subjects $\leq 1000$) and domain diversity (recorded only in indoor conditions) of these two previous competitions were still limited for evaluating the performance of FAS methods 'in the wild'. Recently, a large-scale FAS dataset, namely CelebA-Spoof~\cite{zhang2020celeba}, containing 625,537 face images of 10,177 subjects, was released. It is still the largest publicly available face PAD dataset in terms of the number of images and the subjects. Leveraging the CelebA-Spoof dataset, the CelebA-Spoof Challenge on Face Anti-Spoofing was organized in conjunction with the European Conference on Computer Vision (ECCV) 2020 Workshop on Sensing, Understanding and Synthesizing Humans\footnote{\url{https://sense-human.github.io/}}. The goal of this challenge was to boost the research on large-scale face anti-spoofing. 

The ECCV2020 challenge was also hosted on the CodaLab platform\footnote{\url{https://competitions.codalab.org/competitions/26210}}. After registering to the competition, each team was allowed to submit their models to the Amazon Web Services (AWS) and allocated with one 16 GB Tesla V100 GPU to perform online evaluation on the hidden test set. The encrypted prediction files, including the results for each data sample in hidden test set, were sent to the teams via an automatically generated email after their requested online evaluation was ready. The teams were required to upload their encrypted prediction files to the CodaLab platform for ranking the algorithms. 

The ECCV2020 challenge lasted for nine weeks from August 28, 2020 to October 31, 2020. During the contest, the participants had access to the public CelebA-Spoof dataset and were restricted to use only the public CelebA-Spoof training dataset for building their models. The results of the challenge were announced on February 10, 2021. A total number of 134 participants registered for the competition, while 19 teams made valid submissions in the end. The summarized information and results of top five teams are shown in Table~\ref{tab:celebA}. It is surprising to see that all top three teams were from industry, indicating even increasing attention for PAD in real-world AFR applications. It is worth noting that the top three teams achieved TPR=100$\%@FPR=5*10^{-3}$, indicating the effectiveness of the solutions for large-scale face PAD on the CelebA-Spoof dataset.


\begin{table}[!h]
\centering
\caption{ Final results of the top-5 teams in the Challenge ECCV2020 on FAS.\label{tab:celebA}}
{

\scalebox{0.9}{
\begin{tabular}{clcccccccc}
\toprule
Ranking & Team   &  User   & Affiliation  &\begin{tabular}[c]{@{}c@{}}$TPR $\\ $@FPR=10^{-3}$ \end{tabular}   & \begin{tabular}[c]{@{}c@{}}$TPR $ \\ $@FPR=5*10^{-3}$ \end{tabular}   & \begin{tabular}[c]{@{}c@{}}$TPR$ \\ $@FPR=10^{-6}$ \end{tabular} \\ \midrule

1 & ZOLOZ
& ZOLOZ & ZOLOZ
& 1.00000  & 1.00000 & \textbf{1.00000}     \\ \midrule

2 & MM
& liujeff & Meituan
& 1.00000  & 1.00000 & 0.99991     \\ \midrule

3 & AFO
& winboyer & Meituan
& 1.00000  & 1.00000 & 0.99918     \\ \midrule

4 & k\_
& k\_ & -
& 0.99973  & 0.99927 & 0.98026     \\ \midrule

5 & SmartQ
& SmartQ & -
& 0.99963  & 0.99872 & 0.96938               \\ \bottomrule
\end{tabular}
}
}{}
\end{table}

\begin{figure}[h]
	\begin{center}
	\includegraphics[width=1.0\linewidth]{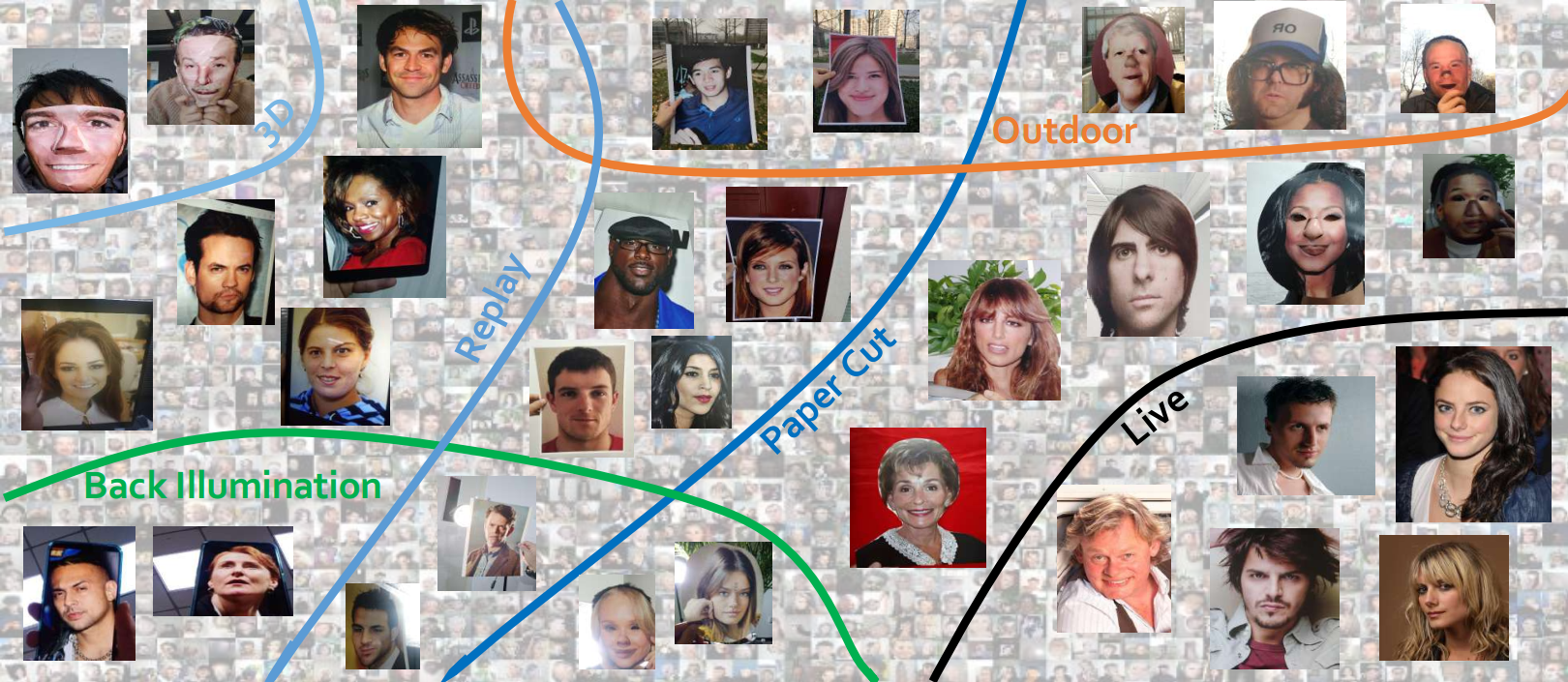}
	\end{center}
	\caption{Representative samples and attributes in the CelebA-Spoof dataset~\cite{zhang2020celeba}.}
	\label{samp_data}
\end{figure}

\subsubsection{Dataset}

The ECCV2020 challenge employed the CelebA-Spoof dataset~\cite{zhang2020celeba} for training and evaluation purposes. The CelebA-Spoof is a large-scale face PAD dataset that has 625,537 images corresponding to 10,177 subjects, including 43 rich attributes on face, illumination, environment and PA types. Bona fide facial images were selected from the CelebA dataset~\cite{liu2015deep} but they were also manually examined to find and remove possible "attack" samples, including posters, advertisements and artistic drawings. The corresponding attack samples with four different PAIs (\ie, 2D Print, cut paper, video-replay and 3D paper mask) were collected and annotated to form the CelebA-Spoof database. Among the 43 rich attributes, 40 attributes describe the bona fide images, including all facial components and accessories (\eg, skin, nose, eye, eyebrow, lip, hair, hat and eyeglasses), while the three remaining attributes describe the attack samples, including PAI, environments (\eg, indoor and outdoor) and illumination conditions (\eg, strong, weak, back, dark and normal). Some typical samples in the CelebA-Spoof database are shown in Fig.~\ref{samp_data}.

\subsubsection{Evaluation protocol and metrics}

The training, validation and test sets of the original CelebA-Spoof database were split into subject-disjoint folds with a ratio of 8 : 1 : 1. The compilation of the hidden test devised for the ECCV2020 challenge was as the same as for the public test set. All the teams participating the competitions were restricted to train their algorithms only using the training subset of the publicly available CelebA-Spoof dataset, thus the use of both proprietary and public external datasets was explicitly forbidden. 

Unlike in the two previous contests (\ie,  CVPR2019~\cite{ChallengeCVPR2019} and CVPR2020~\cite{cvpr2020challenge}), FPR@TPR based evaluation criteria were adopted for the ECCV2020 challenge. The TPR$@$FPR=5$^{-3}$ determined the final ranking but also TPR@FPR=10$^{-3}$ and TPR@FPR=10$^{-4}$ values were reported. In the case if the TPR@FPR=5$^{-3}$ for two submitted algorithms were the same, the one with higher TPR@FPR=10$^{-4}$ would rank better. 

\begin{table*}[h]
\caption{Summary of the top-ranked solutions~\cite{zhang2021celeba} in the ECCV2020 challenge. \label{tab:CelebA_solutions}}
{
\scalebox{0.8}{
\begin{tabular}{clcccccccc}
\toprule
Team   &  Input    &Model   & Ensemble strategy   \\ \midrule

ZOLOZ
& Face
& FOCUS, AENet, ResNet, Attack types, Noise Print  & Heuristic voting strategy  \\ \midrule

MM
& Face + Patches
 &  CDCN++, LGSC, SE-ResNet50, EfficientNet-B7, SE-ResNeXt50 & Weight-after-sorting     \\ \midrule

AFO
& Patches
 &  CDCN, CDC-DAN, SE-ResNeXt26, Light-weighted Network & Weighted summation                \\ \bottomrule
\end{tabular}
}
}{}
\end{table*}

\subsubsection{Results and Discussion}

In this section, we first analyze the top three solutions as well as their results and then we discuss the algorithms and challenge in general.

\textbf{Analysis on the top three solutions:}~Table~\ref{tab:CelebA_solutions} summarizes the FAS solutions of the top three teams. It is not surprising that all the best-performing solutions exploited ensembles of multiple deep models to achieve more robust performance. As the use of external training data in addition to the competition dataset was explicitly forbidden, the features of a single model can easily overfit and learn specific attack cues, while stacking the feature representations of multiple (deep) models can be more generalizing to alleviate this issue. To be more specific, the two best teams ZOLOZ and MM utilized more advanced ensembling strategies compared with the team AFO considering only straightforward weighted summation. The team ZOLOZ proposed a heuristic voting scheme at the score level to form robust combinations of different models, whereas the team MM proposed a novel 'weighting-after-sorting' strategy based on particle swarm optimization (PSO)~\cite{kennedy1995particle} algorithm for their model ensembles. All three teams utilized at least five different deep model architectures, of which some (\eg, FOCUS~\cite{zhang2021celeba}, AENet~\cite{zhang2020celeba}, Noise Print~\cite{jourabloo2018face}, CDCN~\cite{yu2020searching}, CDCN++~\cite{yu2020searching}, LGSC~\cite{feng2020learning}, CDC-DAN~\cite{zhang2021celeba}) were aiming at pixel-wise fine-grained spoof pattern representation, while some others (\eg, ResNet~\cite{he2016deep}, SE-ResNet~\cite{hu2018squeeze}, EfficientNet~\cite{tan2019efficientnet}, ResNeXt~\cite{xie2017aggregated}) focused on extracting semantic cues that could be complementary to improve face PAD performance. As for the model inputs, both whole facial images and local image patches were utilized by team MM, while the team ZOLOZ and AFO considered only the whole facial images and local image patches, respectively. It can be seen from the Table~\ref{tab:celebA} that all top three teams achieved excellent performance reaching TPR$\textgreater$0.999$@FPR=10^{-6}$, indicating the effectiveness of deep multi-model ensembles on the competition data.

\textbf{Discussion:}~Although the aforementioned best solutions achieved very promising results on the CelebA-Spoof dataset, there were still some shortcomings with the ECCV2020 challenge. The hidden test set is rather similar compared with the training data because the CelebA-Spoof dataset was simply divided into subject-disjoint training, development and test folds, thus not explicitly taking into account specific known issues related to domain generalization or unknown PAs. Furthermore, there were no restrictions on the size or number of deep models, which was also disappointing from real-world deployment point of view. Finally, compared with previous two face PAD challenges (\ie, CVPR2019~\cite{ChallengeCVPR2019} and CVPR2020~\cite{liu2021casia}), detailed ablation studies (\eg, impact of each sub-model prior stacking) were missing, as well as the source codes of the solutions were not made public available in the ECCV2020 challenge, thus limiting the transparency and, consequently, usefulness of the whole competition to the FAS community.

\subsection{LivDet-Face 2021 - Face Liveness Detection Competition (IJCB2021)}
\label{sec:com4}


Recent literature surveys (\eg,~\cite{ming2020survey,yu2021deep}) have concluded that both handcrafted and deep features yield in satisfying classification performance in identifying known PAIs but often fail to detect unknown PAIs and more sophisticated facial artefacts, thus continuous efforts are necessary to update face anti-spoofing algorithms to detect rapidly evolving PAs. Although earlier the CVPR2020 cross-ethnicity face PAD challenge considered also a cross-PAI setting (\ie, training on the video-replay attacks and testing on the print and mask PAIs), the types and quality of the unknown PAIs were still limited from the generalized PAD point of view. To address this issue, the LivDet-Face competition\footnote{\url{https://face2021.livdet.org/}} was organized in conjunction with IJCB2021.

The registration of the IJCB2021 LivDet-Face competition began on February 15, 2021 and ended on April 25, 2021, while the final submission deadline was April 30, 2021. The objective of the competition was to evaluate the performance of the state-of-the-art facial PAD algorithms against traditional and novel PAIs. The competition had two separate tracks for image and video data, and the competitors were allowed to participate both tracks. Different from all previous competitions, IJCB2021 LivDet-Face contest did not provide any specific training dataset to the competitors, thus the competitors were free to use any proprietary and/or publicly available data to train their algorithms, replicating more realistic and challenging practical AFR application scenarios.

Both academic and industrial organizations were welcome to participate in IJCB2021 LivDet-Face competition anonymously or non-anonymously. In total thirty international teams registered to the competition, including ten submissions for the image track and six submissions for the video track. Finally, six submission could be successfully tested by the organizers for the image track and five submission for the video track. Unsuccessful tests were due to software issues, which were communicated with the participants.

\begin{figure}[h]
	\begin{center}
	\includegraphics[width=1.0\linewidth]{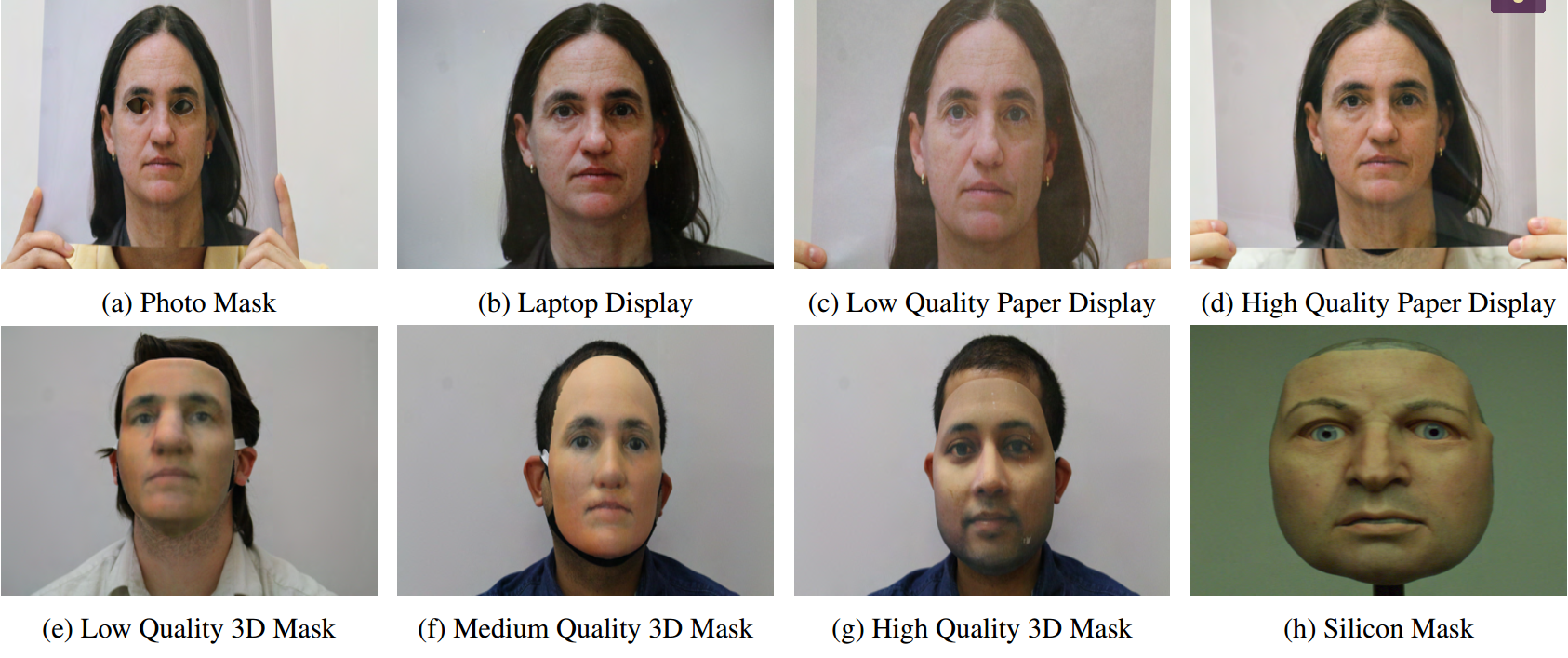}
	\end{center}
	\caption{Samples of each PA type present in the IJCB2021 LivDet-Face test set~\cite{purnapatra2021face}.}
	\label{fig:LivDet}
\end{figure}

\subsubsection{Dataset}

No official training dataset was shared by the organizers of the IJCB2021 LivDet-Face competition. Instead, participants were encouraged to use any data available to them (\ie, from both public and proprietary sources) to train and tune their algorithms. The organizers shared only few (no more than two) examples of the known PAIs to familiarize the competitors with the test dataset, while the remaining samples of the disclosed PAI types were considered as unknown to the competitors. The test dataset used in the IJCB2021 LivDet-Face competition was a combination of the data from two of the organizing institutions: Clarkson University (CU) and Idiap Research Institute. The data consisted of 724 images (135 bona fide and 589 PAI samples) and 814 videos (125 bona fide and 689 PAI samples) for the image and video tracks, respectively, which were collected using altogether five different sensors (digital single-lens reflex camera (DSLR), iPhone X, Samsung Galaxy S9, Google Pixel and Basler aA1920-150uc) from in total 48 live subjects. The video lengths of the test dataset were up to six seconds. Eight PAIs for the image track and nine PAIs for the video track were included in the dataset (see Table~\ref{tab:LivDet} for a summary and Fig.~\ref{fig:LivDet} for typical examples for each PAI type).

Regarding the 2D PAIs, 100 low-quality (LQ) print paper attacks, 100 high-quality (HQ) photo paper attacks, 100 static display (SD) and video-replay (VR) attacks on laptop screen, and 100 2D photo mask attacks were collected from 25 live subjects using four different sensors. Specifically, the 100 video-replay attacks were used as an unknown PAI for the video category of the competition and introduced in the few validation samples that were shared with the competitors. In addition to the 2D attacks, three different qualities of 3D masks (low, medium and high) as well as the silicon mask attacks were included in the test dataset for the both competition tracks. A total number of 24 images and 24 videos of low-quality (LQ) 3D masks were created corresponding to six live subjects. Also, 12 medium-quality (MQ) 3D mask and 12 high-quality (HQ) images/videos recorded corresponding to three live subjects were included in the test dataset. The HQ 3D masks were kept as an unknown PAIs for the competitors and the test dataset. In total 141 image/video samples of the wearable 3D silicon masks were collected using five different sensors.

\begin{table*}[t]
\centering
\caption{Summary of the test dataset used in the IJCB2021 LivDet-Face~\cite{purnapatra2021face}. \label{tab:LivDet}}
{
\scalebox{0.75}{
\begin{tabular}{clcccccccc}
\toprule
Class   &  Types of PAIs    & Images   & Videos  & Sensors   \\ \midrule

Live
& -
& 135  & 125  & DSLR, iPhone X, Samsung Galaxy S9, Google Pixel \\ \midrule

PAI
& Laptop Display (DL)
& 100  & 100  & DSLR, iPhone X, Samsung Galaxy S9, Google Pixel \\ \midrule

PAI
& Photo Mask (PM)
& 100  & 100  & DSLR, iPhone X, Samsung Galaxy S9, Google Pixel \\ \midrule

PAI
& Low-Quality Paper Display
& 100  & 100  & DSLR, iPhone X, Samsung Galaxy S9, Google Pixel \\ \midrule

PAI
& High-Quality Paper Display
& 100  & 100  & DSLR, iPhone X, Samsung Galaxy S9, Google Pixel \\ \midrule

PAI
& Low-Quality 3D Mask
& 24  & 24  & DSLR, iPhone X, Samsung Galaxy S9, Google Pixel \\ \midrule

PAI
& Medium-Quality 3D Mask
& 12  & 12  & DSLR, iPhone X, Samsung Galaxy S9, Google Pixel\\ \midrule

PAI
& High-Quality 3D Mask
& 12  & 12  & DSLR, iPhone X, Samsung Galaxy S9, Google Pixel \\ \midrule

PAI
& Silicon Mask
& 141  & 141  & DSLR, iPhone X, Samsung Galaxy S9, Google Pixel, Basler acA1920-150uc \\ \midrule

PAI
& Video Display (VD)
& -  & 100  & DSLR, iPhone X, Samsung Galaxy S9, Google Pixel \\ \bottomrule

\end{tabular}
}
}{}
\end{table*}

\subsubsection{Evaluation protocol and metrics}

During the IJCB2021 LivDet-Face competition, at least two samples of majority of the considered PAIs (except high-quality 3D masks and video-replay attacks) were shared with the competitors of both image and video tracks as a small validation set to fine-tune their algorithms. The performance of an algorithm for each sample was determined by a output ("liveness") score ranging between 0 to 100 with a threshold of 50, while the score of 1000 indicates undetected samples. The test samples with scores less than 50 were classified as PA, whereas the scores of 50 and above were classified as bona fide. Most of the competitors normalized their output scores at their end and provided a score of 0, 100 or 1000 (if undetected) based on their classification output. If the submitted algorithms provided a score of 1000 for the PAs, the result was considered as a correct decision as the algorithm was able to reject an attack, thus is not included in attack presentation classification errors. A score of 1000 for bona fide samples were considered as incorrect, thus was accumulated in bona fide classification errors.

Following the recommendations of the ISO/IEC 30107-3~\cite{ACER} standard, APCER, BPCER, and ACER (see, Section \ref{sec:metrics}) were used as evaluation metrics like in the CVPR2020 challenge. Since all algorithms were required to deliver normalized liveness scores in the range of 0-100, $t$=50 was used as the decision threshold to calculate APCER and BPCER. The final ranking of teams was based on the ACER calculated over all the test samples.

\begin{table*}[h]
\caption{Summary of the solutions of in IJCB2021 LivDet-Face~\cite{purnapatra2021face}. \label{tab:LivDet_solutions}}
{
\scalebox{0.8}{
\begin{tabular}{clcccccccc}
\toprule
Team   &  Training/validation    &Model   & Ensemble strategy   \\ \midrule

Fraunhofer IGD
& CRMA for validation
& 12 models (DeepPixBis, ResNeXt, etc.)  &  FDR weights  \\ \midrule

SiMiT Lab
& \begin{tabular}[c]{@{}c@{}}Replay Mobile, SiW, Oulu-NPU,\\ 3DMAD for training \end{tabular}
 &  DeepPixBis, EfficientNet-B7 & Mean score fusion     \\ \midrule
 
CLFM
& -
 &  CDCN & -                \\ \midrule

FaceMe
& -
 &  3 models (DepthNet, Digital signal processor, etc.)  & -     \\ \midrule

little tiger
& Glint360k for pre-training
 &  5 models (ResNet50, ResNext26, CDCN, etc.)  & -     \\ \midrule

NTNU Gjøvik
& \begin{tabular}[c]{@{}c@{}}SWAN, CASIA-FASD, NTNU-\\Silicon Mask for training\end{tabular}
 & \begin{tabular}[c]{@{}c@{}} 6 models (Resnet18, Resnet50,  InceptionV3, 9\\VGG1, VGG16, Alexnet) with 2 linear SVM \end{tabular} & Majority voting               \\ \bottomrule
\end{tabular}
}
}{}
\end{table*}

\subsubsection{Results and Discussion}

In this section, we first analyze the solutions of both tracks. Then, the result analysis of the image and video tracks is presented. Finally, we discuss the algorithms and the challenge in general.

\textbf{Analysis on the solutions:}~Table~\ref{tab:LivDet_solutions} summarizes the FAS solutions of the six teams. The teams SiMiT Lab and NTNU Gjøvik utilized several public datasets for training, including the 2D PAIs (\eg, Replay-Mobile~\cite{Costa2016The}, SiW~\cite{Liu2018Learning}, Oulu-NPU~\cite{Boulkenafet2017OULU}, SWAN~\cite{ramachandra2019smartphone}, CASIA-FASD~\cite{Zhang2012A}) and 3D mask attacks (\eg, 3DMAD~\cite{erdogmus2014spoofing} and NTNU-Silicon Mask~\cite{ramachandra2019custom}) datasets. The team Fraunhofer IGD improved the generalization of their algorithm by using the 50 attacks and bona fide samples from the Real Mask Attack Database (CRMA)~\cite{fang2021real} as unknown development data to tune the decision threshold.

Like in the previous three challenges discussed already in this chapter, most of the solutions in IJCB2021 LivDet-Face competition used ensembles of multiple deep models to achieve more robust performance. The teams Fraunhofer IGD, SiMiT Lab, FaceMe, little tiger and NTNU Gjøvik stacked 12, 2, 3, 5 and 6 models in their submitted systems to make the final PAD decision. Three kinds of models were considered in these solutions: 1) hand-crafted models with digital signal processing; 2) pixel-wise supervised models (\eg, DeepPixBis~\cite{george2019deep}, DepthNet~\cite{Liu2018Learning} and CDCN~\cite{yu2020fas}); 3) binary cross-entropy supervised models (\eg, ResNet~\cite{he2016deep}, ResNeXt~\cite{xie2017aggregated}, VGG~\cite{simonyan2014very}, Inception~\cite{szegedy2016rethinking} and Alexnet~\cite{Krizhevsky2012ImageNet}). The team Fraunhofer IGD adopted Fisher-discriminative ratio (FDR) weights~\cite{damer2014biometric} for 12 models to get a combined face PAD decision, while the team NTNU Gjøvik considered a simple majority voting strategy to make decision from six models, and the team SiMiT Lab fused the scores from two models with shuffled patch-wise supervision~\cite{kantarci2021shuffled}. Some competitors (\eg, teams CLFM and FaceMe) were lacking these kinds of details in their method descriptions, thus making it impossible to draw conclusions on some factors in our result analysis, including data fusion approaches and training/tuning data.

\begin{table}[]
\caption{Results of the image track in the IJCB2021 LivDet-Face competition. The APCER is respectively calculated for each type of PAI, and then averaged for final ACER calculation. \label{tab:LivDet_image}}
\centering
{
\scalebox{0.9}{
\begin{tabular}{@{}c|cc|c|c|cccc|cc|c@{}}
\toprule

\multirow{2}{*}{Team} & \multicolumn{2}{c|}{Paper}  & \multirow{2}{*}{Replay} & \multirow{2}{*}{2D Mask} & \multicolumn{4}{c|}{3D Mask} & \multirow{2}{*}{BPCER(\%)} & \multirow{2}{*}{ACER(\%)} & \multirow{2}{*}{Ranking}\\
\cmidrule(lr){2-3}
\cmidrule(l){6-9}
& LQ & HQ & & & LQ & MQ & HQ & Silicon & & &

\\ \midrule
Fraunhofer IGD
& \textbf{0}  & 24   & 45  & 14.7  & 4.17
& 8.33  & \textbf{14.29}  & \textbf{16.31} & \textbf{15.33} & \textbf{16.47}  & 1 

\\ \midrule
CLFM
& 6.06  & \textbf{10}   & 8  & \textbf{5.88}  & \textbf{0}
& 16.67  & 21.43  & 34.75 & 24.08 & 18.71  & 2 

\\ \midrule
FaceMe
& 22.22  & 11   & \textbf{3}  & 11.76  & 66.67
& 66.66  & 50  & 57.45 & 16.06 & 20.72  & 3  

\\ \midrule
little tiger
& 41.41  & 52  & 4  & 58.82  & 54.17
& 25  & 28.57  & 82.98 & 21.17 & 33.92  & 4  

\\ \midrule
SiMiT Lab
& 7.07  & 18   & 43  & 15.68 & 16.66
& \textbf{0} & 42.85  & 80.85 & 51.09 & 42.05  & 5  

\\ \midrule
Anonymous
& 78.78  & 86   & 77  & 89.21  & 87.5
& 83.33  & 100  & 98.58 & 16.79 & 49.35 & 6   \\

\bottomrule
\end{tabular}
}
}{}
\end{table}

\textbf{Results of the image track:}~Table~\ref{tab:LivDet_image} summarizes the results of the image track. The team Fraunhofer IGD was the winner obtaining the lowest ACER = 16.47\%, followed by the team CLFM with a narrow margin in ACER = 18.71\%. The winning team Fraunhofer IGD achieved the lowest BPCER = 15.33\% among the six competitors. The six competitors achieved highly varying performances across the different PAI types. The algorithm submitted by the team Fraunhofer IGD detected all the low-quality paper display attacks and the algorithm by CLFM successfully detected all the low-quality 3D mask samples. Team CLFM’s algorithm also performed the best with APCER = 10\% for high-quality photo paper display samples but achieved an unsatisfying BPCER = 24.08\%. Similarly, the team FaceMe, who achieved third place in the image track achieved the best APCER = 3\% for laptop display samples and achieved BPCER = 16.06\%. The team SiMiT Lab successfully detected all the medium-quality 3D mask samples with APCER = 0\%, which was best among all the competitors, but achieved also the worst BPCER = 51.09\%.

Comparing the performance of the top two ranked solutions in the image track, it is obvious that the models performed better against low-quality PAIs than higher quality PAIs. The team Fraunhofer IGD obtained APCER = 0\% for the low-quality paper display and APCER = 24\% for high-quality paper display. Similarly, the team CLFM obtained APCER = 6.06\% for low-quality paper display, and APCER = 10\% for high-quality paper display. The same trend can be observed for the different quality of 3D face masks. The team Fraunhofer IGD obtained APCER = 4.17\% for low-quality 3D masks compared with APCER = 8.33\% for medium-quality 3D masks, APCER = 14.29\% for high-quality 3D masks and APCER = 16.31\% for high-quality silicon masks. Similarly, the team CLFM achieved APCER = 0\% against low-quality 3D masks compared with APCER = 16.67\% for medium-quality 3D masks, APCER = 21.43\% for high-quality 3D masks and APCER = 34.75\% for high-quality silicon masks.

\begin{table}[]
\caption{Results of the video track in the IJCB2021 LivDet-Face competition. \label{tab:LivDet_video}}
\centering
{
\scalebox{0.9}{
\begin{tabular}{@{}c|cc|cc|c|cccc|cc|c@{}}
\toprule

\multirow{2}{*}{Team} & \multicolumn{2}{c|}{Paper}  & \multicolumn{2}{c|}{Replay}& \multirow{2}{*}{2D Mask} & \multicolumn{4}{c|}{3D Mask} & \multirow{2}{*}{BPCER(\%)} & \multirow{2}{*}{ACER(\%)} 
& \multirow{2}{*}{Ranking}\\
\cmidrule(lr){2-3}
\cmidrule(lr){4-5}
\cmidrule(l){7-10}
& LQ & HQ & SD & VR & & LQ & MQ & HQ & Silicon & & 

\\ \midrule
FaceMe
& 8  & 10.10   & 18  & 16  & 6.93
& 40  & 45.45  & 38.46 & 9.22 & 14.29 & \textbf{13.81} & 1 

\\ \midrule
Fraunhofer IGD
& \textbf{1}   & 25.25   & 29  & 9  & 1
& 4  & 9.09  & \textbf{0} & 12.77 & 16.67 & 14.49 & 2

\\ \midrule
CLFM
& 4  & \textbf{4.04}   & \textbf{8}  & \textbf{1}  & \textbf{0}
& \textbf{0}  & 27.27 & 7.69 & \textbf{1.42} & 39.68 & 21.49 & 3  

\\ \midrule
NTNU Gjøvik-V1
& 50  & 59.60  & 83  & 75  & 18.81
& 36  & 18.18  & 46.15 & 21.28 & \textbf{4.76} & 26.51 & 4 

\\ \midrule
NTNU Gjøvik-V2
& 5  & 9.09   & 32  & 20 & 1
& \textbf{0} & \textbf{0}  & \textbf{0} & 33.33 & 51.59 & 34.05 & 5
\\

\bottomrule
\end{tabular}
}
}{}
\end{table}

\textbf{Results of the video track:}~Table~\ref{tab:LivDet_video} summarizes the results of the five solutions in the video track of the IJCB2021 LivDet-Face competition. The team FaceMe was the winner with the ACER = 13.81\% followed by the team Fraunhofer IGD with a narrow margin in ACER = 14.49\%. The lowest BPCER = 4.76\% was achieved by the team NTNU Gjøvik. The team CLFM performed well in detecting paper and video-replay attacks (with APCER $\textless$3.30\% for all scearios) but ranked third due to the bad BPCER performance (39.68\%). The team NTNU Gjøvik-V2 performed well against 3D face mask attacks achieving APCER = 0\% for the three different types of 3D masks.

It can be observed that the top two solutions performed better against low-quality PAIs compared to higher-quality PAIs. For example, the performance of the team Fraunhofer IGD against low-quality paper display was APCER = 1\%, while APCER = 25.25\% was obtained against the high-quality paper display attacks. The same trend can be observed for the different quality of 3D face masks as well. The performance of the team Fraunhofer IGD against low-quality 3D masks was APCER = 4\%, which was obviously better than that of the medium-quality (APCER = 9.09\%) and high-quality silicon masks (APCER = 34.75\%). However, the performance of high-quality 3D masks was better than that of any other 3D mask category with APCER = 0\%.

\textbf{Discussion:}~Compared with the three earlier competitions~\cite{liu2019multi,cvpr2020challenge,zhang2021celeba} introduced in this chapter, a significant degradation in the overall performance can be observed. This can be due to several factors, such as: 1) increased complexity in the test dataset with nine different PAI types; 2) introduction of three novel attack types with limited availability, or not covered at all, in the public datasets; 3) lack of specific competition training dataset, \ie, choice of training data up to the competitors; 4) domain shifts between the training and test conditions in terms of environmental factors, sensors, quality of PAIs, and the introduction of unknown PAIs. The results from this competition indicate that generalized face PAD is still far away from a solved research problem. 

Despite its important findings on the current state of face PAD 'in the wild', the IJCB2021 LivDet-Face competition still had some shortcomings. First, as there were no pre-defined training sets and, consequently, no restrictions on the diversity and scale of the private or public datasets, it is unfair to evaluate the performance of the different approaches because it is impossible to explore the hidden reasons behind the differences in performance and to tell if an algorithm actually better than another one, or is it in fact a matter of the amount and quality of training data. Second, no ablation studies or open-source code for most of the solutions were provided, which again limits the transparency and, consequently, usefulness of the results and findings to the FAS community. Finally, as the complexity of the proposed systems was not limited, mainly huge ensembles of deep models were adopted by the participants, while none of the teams proposed interesting novel efficient face PAD approaches.

\subsection{3D High-Fidelity Mask Face Presentation Attack Detection Challenge (ICCV2021)}
\label{sec:com5}


As seen in two recent competitions (\ie, CVPR2020~\cite{cvpr2020challenge} and IJCB2021 LivDet-Face ~\cite{purnapatra2021face}), face PAD performance of state-of-the-art methods drops significantly under unknown 3D mask attacks. However, the previous competition datasets contained in general only a limited number of types and samples of 3D facial masks, thus there is a large gap in between the existing benchmarks and 3D mask attack detection in real-world conditions. To alleviate the threats posed by 3D mask attacks and to improve the reliability of face PAD methods under emerging 3D mask attacks in various scenarios, the 3D High-Fidelity Mask Face Presentation Attack Detection Challenge~\cite{liu20213d} was organized in conjunction with the International Conference on Computer Vision (ICCV) in 2021 using the very recently constructed CASIA-SURF High-Fidelity Mask (HiFiMask) dataset~\cite{liu2021contrastive}. 

 The ICCV2021 challenge was conducted using the CodaLab platform\footnote{\url{https://competitions.codalab.org/competitions/30910}}, attracting $195$ teams from all over the world. A summary with the names and affiliations of teams that entered the final stage of the contest is shown in Table~\ref{table:team-affiliations}. Again, the majority of the final participants came from industrial institutions, and all the six best-performing teams represented companies. This indicates clearly that mask attack detection is no longer limited to academic research but also a crucial problem in real-world AFR applications. The results of the top three teams were far better than the baseline results~\cite{liu2021contrastive}, thus greatly improving the performance of 3D high-fidelity mask attack detection.

\begin{table}[]
\footnotesize
\centering
\caption{Names, affiliations and final ranking of the teams participating in the ICCV2021 challenge.}
\begin{tabular}{|c|c|c|c|}
\hline
Ranking & Team Name                                                    & Leader Name & Affiliation                                                                                      \\ \hline \hline
1       & VisionLabs                                                   & Oleg Grinchuk &  Visionlabs.ai                                                                                  \\ \hline
2       & WeOnlyLookOnce                                               & Ke-Yue Zhang &  Tencent Youtu Lab                                                                               \\ \hline
3       & CLFM                                                         & Samuel Huang &  FaceMe                                                                                          \\ \hline
4       & oldiron666                                                   & Zezheng Wang &  Kuaishou Technology                                 \\ \hline
5       & Reconova AI-LAB   & Mingmu Chen &  Reconova Technology                                 \\ \hline
6       & inspire                                                      & Jiang Hao &  Bytedance Ltd.                                                                                     \\ \hline
7       & Piercing Eyes                                                & Hyokong &  National University of Singapore                        \\ \hline
8       & msxf\_cvas                                                   & Liang Gao, MaShang &  Consumer Finance Co.,Ltd   \\ \hline
9       & VIC\_FACE                                                    & Cheng Zhen &  Meituan                                                                                           \\ \hline
10      & DXM-DI-AI-CV-TEAM & Weitai Hu &  Du Xiaoman Financial                                                                               \\ \hline
11      & fscr                                                         & Artem Petrov &  \begin{tabular}[c]{@{}c@{}}Peter the Great St. \\ Petersburg Polytechnic University\end{tabular} \\ \hline
12      & VIPAI                                                        & Yao Xiao &  Zhejiang University                                                                                 \\ \hline
13      & reconova-ZJU                                                 & Zhishan Li &  Zhejiang University                                                                               \\ \hline
14      & sama\_cmb                                                    & Yifan Chen &  Chinese Merchants Bank(CMB)                            \\ \hline
15      & Super                                                        & Yu He &  Technische Universität München                  \\ \hline
16      & ReadFace                                                     & Zhijun Tong &  ReadFace                                                                                         \\ \hline
17      & LsyL6                                                        & Dongxiao Li &  Zhejiang University                                                                              \\ \hline
18      & HighC                                                        & Minzhe Huang & Akuvox (Xiamen) Networks Co., Ltd.               \\ \hline
\end{tabular}
\label{table:team-affiliations}
\end{table}

\subsubsection{Dataset}
\label{sec:dataset}

\begin{figure}[t]
	\begin{center}
	\includegraphics[width=0.8\linewidth]{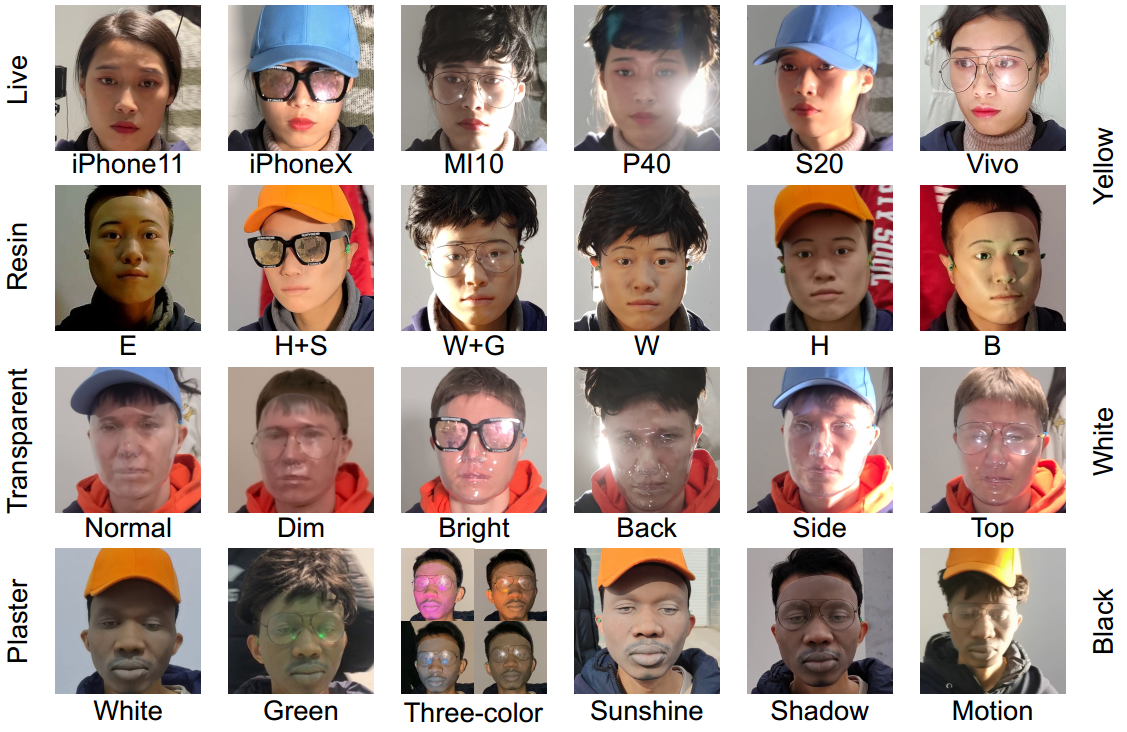}
	\end{center}
	\caption{ Samples from the HiFiMask dataset~\cite{liu2021contrastive}. The first row shows six kinds of imaging sensors. The second row shows six kinds of appendages, of which E, H, S, W, G, and B are the abbreviations for 'empty', 'hat', 'sunglasses', 'wig', 'glasses', and 'messy background', respectively. The third row shows six kinds of illumination conditions, and the fourth row represents six deployment scenarios.}
	\label{samp_3Dmask}
\end{figure}

HiFiMask~\cite{liu2021contrastive} is currently the largest 3D face mask PAD dataset, consisting of $54,600$ videos corresponding to $75$ subjects with three skin tones, including $25$ subjects in yellow, white, and black, respectively. The database contains three high-fidelity masks for each identity, which are made of transparent, plaster and resin materials, respectively. During the acquisition process, six complex scenes were considered for recording the videos (\ie, white light, green light, periodic three-color light, outdoor sunshine, outdoor shadow and motion blur). For each scene, there are six videos captured under different lighting conditions (\ie, normal, dim, bright, back, side and top) to explore the impact of directional lighting. Periodic lighting within [0.7, 4] Hz in the first three scenarios (see, the first three columns of the last row in Fig.~\ref{samp_3Dmask} for examples) tries to mimic the natural human pulse variations to fool remote photoplethysmography (rPPG) based mask detection technology (\eg, ~\cite{li2016generalized}). Finally, seven mainstream imaging devices (\ie, iPhone11, iPhone X, MI10, P40, S20, Vivo and HJIM) are utilized for recording the videos in order to ensure high resolution and imaging quality corresponding to modern mobile devices. The original videos were not provided due to huge amount of data. In order to decrease the size of the dataset, the organizers sampled every tenth frame of each video and applied a fast face detector \cite{zhang2017faceboxes} to remove most of the background information from the sampled video frames, thus the final competition data consisted of coarsely pre-cropped facial images. Some typical samples of pre-processed video frames in the HiFiMask dataset are presented in Fig.~\ref{samp_3Dmask}.

In order to increase the difficulty of the competition and meet the real-world deployment requirements, an 'open-set' test protocol was utilized to comprehensively evaluate the discriminative and generalization power of face PAD algorithms. In other words, the training and development sets contained only subsets of common mask types and operating scenarios, while there were more general mask types and scenarios in the test set. Thus, the distribution of testing sets was more complicated compared to the training and development sets in terms of mask types, scenes, lighting, and imaging devices. Such 'open-set' protocol considers explicitly both 'seen' and 'unseen' domains as well as mask types for evaluation, which is also more valuable from real-world face PAD deployment point of view.
\begin{table}[]
\centering
\caption{Statistical information for the protocols used in the ICCV2021 challenge. Note that 1, 2 and 3 in the third column mean 'transparent', 'plaster' and 'resin' masks, respectively. The numbers in the fourth, fifth, and sixth columns are explained in Section~\ref{sec:dataset}.}

\scalebox{1.0}{
\begin{tabular}{|c|c|c|c|c|c|c|l|l|c|l|l|c|l|l|}
\hline
Subset & Subject & Mask type   & Scene   & Light   & Sensor  & \multicolumn{3}{c|}{\# Live num.} & \multicolumn{3}{c|}{\# Mask num.} & \multicolumn{3}{c|}{\# All num.} \\ \hline
Train  & 45      & 1\&3      & 1\&4\&6    & 1\&3\&4\&6  & 1\&2\&3\&4  & \multicolumn{3}{c|}{1,610}   & \multicolumn{3}{c|}{2,105}   & \multicolumn{3}{c|}{3,715}  \\ \hline
Dev    & 6       & 1\&3      & 1\&4\&6    & 1\&3\&4\&6  & 1\&2\&3\&4  & \multicolumn{3}{c|}{210}     & \multicolumn{3}{c|}{320}     & \multicolumn{3}{c|}{536}    \\ \hline
Test   & 24      & 1$\sim$3 & 1$\sim$6 & 1$\sim$6 & 1$\sim$7 & \multicolumn{3}{c|}{4,335}   & \multicolumn{3}{c|}{13,027}  & \multicolumn{3}{c|}{17,362} \\ \hline
\end{tabular}
}
\label{tab:protocol-3}
\end{table}

As shown in Table~\ref{tab:protocol-3}, every skin tone, part of mask types, such as transparent and resin materials (1, 3), part of scenes, such as white light, outdoor sunshine and motion blur (1, 4, 6), part of lighting conditions, such as normal, bright, back and top (1, 3, 4, 6), and part of imaging devices, such as iPhone 11, iPhone X, MI10, P40 (1, 2, 3, 4) are included in the training and development subsets. All skin tones, mask types, scenes, lighting conditions and imaging devices are present in the test subset. For clarity, the organization of the dataset and quantity of videos for each sub-protocol of the challenge are shown in Table~\ref{tab:protocol-3}.

\subsubsection{Evaluation protocol and Metrics}

The challenge comprised two stages as follows:

\textbf{Development phase:}~(\emph{April. 19, 2021 - June 10, 2021}). During the development phase, the participants had access to the labeled training data and unlabeled development data. The samples in the training set were labeled with the bona fide, two types of masks (1, 3), three types of scenes (1, 4, 6), four kinds of lighting conditions (1, 2, 4, 6) and four imaging sensors (1, 2, 3, 4). The labels of the validation data were not provided to the participants in development phase. Instead, participants could build their models on the labeled training data and then submit their predictions on the development data and receive immediate feedback via the competition leaderboard in the CodaLab platform.

\textbf{Final phase:}~(\emph{June 10, 2021 - June 20, 2021}). During the final phase, the labels for the development set were also made available to the participants and the unlabeled test set was also released. The competitors had to make their predictions on the test samples and upload their solutions to the challenge platform. The organizers did then rerun the best-performing algorithms, and the final ranking of the participants was obtained based from the verified results on the test data. To further facilitate the outcome and findings of the competition to the face PAD community, the best-performing teams were encouraged to make their source code publicly available and provide fact sheets describing their solutions.

\textbf{Evaluation metrics:} Similarly to the previous two competitions (\ie, CVPR2020 and IJCB2021 LivDet-Face~\cite{cvpr2020challenge,purnapatra2021face}), the recently standardized ISO/IEC 30107-3~\cite{ACER} metrics APCER, BPCER, and ACER were selected as the evaluation metrics, and the ACER was the leading evaluation criterion for ranking the submitted systems in the ICCV2021 challenge. The ACER threshold on the test set was determined based on the EER operating point on the development data.

\begin{table*}[]
\caption{Summary of the solutions in the ICCV2021 challenge~\cite{liu20213d}. 'DSFD' and 'DBO' denote dual shot face detector~\cite{li2019dsfd} and deep bilateral operator~\cite{yu2020face}, respectively. \label{tab:3Dmask_solutions}}
{
\scalebox{0.8}{
\begin{tabular}{@{}c|c|c|c|c@{}}
\toprule
Team Name & Pre-processing  & Backbone & Branch & Loss \\ \midrule

VisionLabs
&  DSFD detector  &  EfficientNet-B0  &  6  & BCE \\ \midrule

WeOnlyLookOnce
&  DSFD detector   &  ResNet12  &  2 & 3-class CE with label smoothing \\ \midrule

CLFM
& Crop mouth region & CDCN++  &  1 &  BCE \\ \midrule

Oldiron666
&  Whole face &  Resnet6 (SimSiam)  &  1  & BCE+MSE+Contrastive loss \\ \midrule

Reconova-AI-Lab
&  RetinaFace detector &  ResNet50+YoLoV3-FPN &  3  &  BCE+Focal loss \\ \midrule

inspire
&  RetinaFace detector & SE-ResNeXt101  &  1  & BCE+MSE+Contrastive loss \\ \midrule

Piercing Eye
& Face detection &  CDCN  & 2  &  Depth regression+BCE \\ \midrule

msxf cvas
&  Face detection+alignment &  ResNet34  &  1 &  4-class CE loss \\ \midrule

VIC FACE
&  Face detection &  CDCN with DBO  &  1  & Depth regression \\ \midrule

DXM-DI-AI-CV-TEAM
&  -   &  DepthNet &  1  & BCE with meta learning

\\ \bottomrule
\end{tabular}
}
}{}
\end{table*}

\subsubsection{Results and Discussion}

In this section, we first summarize solutions in the 3D Mask face PAD challenge. Then, we provide our result analysis. Finally, the algorithms and the challenge are discussed in general.

\textbf{Analysis of the solutions:}~Table~\ref{tab:3Dmask_solutions} summarizes the solutions of the most teams participating the ICCV2021 challenge. The source code of the winning team VisionLabs was released\footnote{\url{https://github.com/AlexanderParkin/chalearn_3d_hifi}} while the detailed descriptions as well as the ablation studies of the top three ranked solutions can be found in~\cite{chen2021dual,grinchuk20213d,huang2021single}. Different from all the previous competitions, the ICCV2021 challenge accepted only the results obtained with single (deep) model based systems, thus ensemble strategy with multiple models was explicitly prohibited. As a result, several solutions (\eg, the teams VisionLabs, WeOnlyLookOnce, Reconova-AI-Lab and Piercing Eye) developed multi-branch architectures, which aimed at capturing more diverse PAD-specific feature representations. Also, team VIC FACE proposed a novel architecture in its solution, which integrated the deep bilateral operator in the original CDCN~\cite{ijcai2021} in order to learn more intrinsic features via aggregating multi-level bilateral macro and micro-texture information. Most of the teams adopted DSFD~\cite{li2019dsfd} or RetinaFace~\cite{deng2020retinaface} detector for pre-processing to localize more fine-grained facial region and to filter out partial low-quality face mask attacks. The teams considered face PAD mainly as a binary classification task using BCE loss or as a depth regression problem, but two teams (\ie, WeOnlyLookOnce and msxf cvas) also forced the models to learn more mask type-aware features via fine-grained multiple class cross-entropy (CE) loss.

\begin{table*}[]
\centering
\caption{The final results and team rankings of the ICCV2021 challenge~\cite{liu20213d}. The best results are shown in bold.\label{tab:3dmask_result}}
{
\scalebox{0.93}{
\begin{tabular}{@{}clccccccc@{}}
\toprule
Team Name   & FP   & FN & APCER(\%) & BPCER(\%) & ACER(\%) & Rank \\ \midrule

VisionLabs                                                   
& 492          & \textbf{101} & 3.777          & \textbf{2.330} & \textbf{3.053} & 1

\\ \midrule

WeOnlyLookOnce                                               
& \textbf{242} & 193          & \textbf{1.858} & 4.452          & 3.155 & 2

\\ \midrule

CLFM                                                         
& 483          & 118          & 3.708          & 2.722          & 3.215  & 3

\\ \midrule

oldiron666                                                   
& 644          & 115          & 4.944          & 2.653          & 3.798  & 4

\\ \midrule

Reconova-AI-LAB                                              
& 277          & 276          & 2.126          & 6.367          & 4.247 & 5

\\ \midrule

inspire                                                      
& 760          & 176          & 5.834          & 4.060          & 4.947 & 6

\\ \midrule

Piercing Eyes                                       
& 887          & 143          & 6.809          & 3.299          & 5.054  & 7

\\ \midrule

msxf\_cvas                                                   
& 752          & 232          & 5.773          & 5.352          & 5.562 & 8

\\ \midrule

VIC\_FACE                                                    
& 1152         & 104          & 8.843          & 2.399          & 5.621  & 9

\\ \midrule

DXM-DI-AI-CV-TEAM
& 1100         & 181          & 8.444          & 4.175          & 6.310  & 10

\\ \midrule

fscr                                                         
& 794          & 326          & 6.095          & 7.520          & 6.808 & 11

\\ \midrule

VIPAI                                                        
& 1038         & 268          & 7.968          & 6.182          & 7.075 & 12

\\ \midrule

reconova-ZJU                                                 
& 1330         & 183          & 10.210         & 4.221          & 7.216  & 13

\\ \midrule

sama\_cmb                                                    
& 1549         & 188          & 11.891         & 4.337          & 8.114 & 14

\\ \midrule

Super                                                        
& 780          & 454          & 5.988          & 10.473         & 8.230  & 15

\\ \midrule

ReadFace                                                     
& 1556         & 202          & 11.944         & 4.660          & 8.302  & 16

\\ \midrule

LsyL6                                                        
& 2031         & 138          & 15.591         & 3.183          & 9.387  & 17

\\ \midrule

HighC                                                        
& 1656         & 340          & 12.712         & 7.843  & 18

 \\ \bottomrule
\end{tabular}
}
}{}
\end{table*}

\textbf{Result analysis:}~
The results and ranking of the top 18 teams are shown in Table~\ref{tab:3dmask_result}. The ACER performance of the top three teams were relatively close (within $\textless$3.3\%). One major reason behind this is that the original pre-processed images in the HiFiMask dataset correspond to very coarsely cropped facial regions, including also outliers (\ie, non-facial samples), due to the limited accuracy of the used efficient face detector \cite{zhang2017faceboxes}. Thus, the additional pre-processing with DSFD and mouth region cropping was very beneficial in removing the outlier samples affecting negatively the training of deep models and focusing on the details in the actual facial information discriminating attacks from bona fide samples. The team VisionLabs achieved the best BPCER=2.33\% with only 101 FN samples, while WeOnlyLookOnce had the lowest APCER=1.858\% with 242 FP samples, indicating that the features from multiple facial region based branches and vanilla/CDC branches benefited the spoof cue representations. Moreover, the ACER performance of all the teams were evenly distributed ranging from $3\%$ to $10\%$, which not only indicates the rationality and selectivity of the 3D mask face PAD challenge but also demonstrated the value of the HiFiMask dataset for future face PAD studies.

\textbf{Discussion:}~Some general observations on 3D mask face attack detection can be concluded based on findings of the ICCV2021 competition: 1) the accurate facial localization and sub-region information was very beneficial, focusing on the actual discriminative local facial details and avoiding learning and extraction of irrelevant features for face PAD; and 2) multi-branch-based feature learning was a widely used framework by the participating teams, which benefits the shared low-level feature learning, capturing diverse separated multi-branch features for generalized 3D mask description. However, there were still some shortcomings in the ICCV2021 challenge. The fidelity and the collection of the 3D mask types were still limited. For instance, the face mask in the fourth row and second column of Fig.~\ref{samp_3Dmask} has a very artificial appearance, while also more 3D attack types, such as paper and silicone masks, and wax faces should be considered. Furthermore, only every tenth frame for each video was provided. Such a low frame rate makes it impossible to recover facial physiological signals and, consequently, to study rPPG based 3D mask detection~\cite{liu2021multi,yu2021transrppg}, for instance.

\section{Discussion} 
\label{sec:discussion}

All the recent five competitions were successful in consolidating and benchmarking the current state of the art in face PAD. In the following, we provide general observations and further discussion on the lessons learnt and potential future challenges.

\subsection{General Observations}

It is apparent that the used datasets and evaluation protocols, and also the recent advances in the state of the art reflect the face PAD scheme trends seen in the different contests. The algorithms proposed in the context of the first two multi-modal competitions (\ie, CVPR2019~\cite{cvpr2019challenge} and CVPR2020~\cite{cvpr2020challenge} challenges) exploited the evident visual cues that we humans can observe in the multi-modal imaging data of the CASIA-SURF and CASIA-SURF CeFA datasets, including plain/structural discrepancies in depth modality, material surface reflection differences in NIR modality and natural human movement in dynamic modality. Most solutions adopted exhausted model and hyperparameter selection strategy for each modality and ensembling the best combination of different modalities and models for robust PAD performance in multi-modal setting, \ie, TPR=99.8739\%@FPR=10e-4 on CASIA-SURF and ACER=1.02\% on CASIA-SURF CeFA. Based on these two competitions, it can be seen that the performance gains mostly relied on more powerful models and ensembling strategies, while the essence of multi-modal fusion was not truly explored, which gave, disappointingly, limited insight to multi-modal face PAD community. Therefore, it is necessary to rethink these kinds of multi-modal PAD evaluations especially from the efficient fusion point of view in future multi-modal benchmarks.


In the latest two competitions (\ie, IJCB2021 LivDet-Face~\cite{purnapatra2021face} and ICCV2021 \cite{liu20213d}) on unimodal colour camera based face PAD, generalized and intrinsic bona fide/attack cues and motion analysis were hardly used, whereas the proposed RGB data based features were overfitting in the training data, thus generalized poorly on the test sets. One reason to unsatisfactory performance is that the test sets included more unseen high-fidelity 3D mask types (\eg, plaster masks in ICCV2021 \cite{liu20213d} and high-quality 3D masks in IJCB2021 LivDet-Face~\cite{purnapatra2021face}). Although it was nice to see a diverse set of advanced deep learning based systems and further improved versions of the provided baseline method, it was bit disappointing that entirely novel generalized face PAD solutions for zero-shot unseen attack (especially against 3D masks) detection were not proposed. The best performance in IJCB2021 LivDet-Face (ACER=16.47\%/13.81\% for image/video tracks) and ICCV2021 (ACER=3.053\%) competitions were still limited considering the needs of real-world use cases. As seen in the CVPR2020 competition~\cite{cvpr2020challenge}, the major issues with domain shift and unseen PAs could be at least partially alleviated by introducing depth and NIR sensors as the additional depth and NIR modalities would provide additional shape information for 2D PAI detection and material reflection information to discriminate realistic 3D masks among other facial artefacts from genuine human skin.

Unlike with the first three competitions (\ie, IJCB2011~\cite{chakka2011competition}, ICB2013~\cite{chingovska20132nd} and IJCB2017~\cite{boulkenafet2017competition}), it is interesting to observe that most of the participants came from the industry in the competitions organized from 2019. For example, the majority of the final participants (10
out of 13) of the CVPR2019 challenge~\cite{cvpr2019challenge} and (8 out of 11) of the unimodal track of the CVPR2020 challenge~\cite{cvpr2020challenge} came from industrial institutions. Also, in the most recent ICCV2021 challenge~\cite{liu20213d}, the majority (13 out of 18) of the final participants came from companies, and the six top-performing teams were all from industry, indicating the steadily growing interest and need for practical and reliable face PAD solutions in commercial AFR products.

\subsection{Lessons Learnt}

The competitions have given valuable lessons on designing databases and test protocols, and competitions in general. In the CVPR2019~\cite{cvpr2019challenge} and ECCV2020~\cite{zhang2021celeba}) challenges, the best performance on test data reached TPR=99.8739\%@ FPR=10e-4 and TPR=100\%@FPR=10e-6, respectively. However, the problem of face PAD has not been solved as the error rates (ACER=16.47\%/13.81\%) of the more recent IJCB2021 LivDet-Face reveal that the state of the art in face anti-spoofing still suffers from significant generalization issues in unknown operating conditions. Therefore, we should also rethink the evaluation protocol design in competitions like the CVPR2019~\cite{cvpr2019challenge} and ECCV2020~\cite{zhang2021celeba}) challenges. In the CVPR2019 challenge, the provided validation data had similar distribution with those in test set and the validation set was also allowed to be included in training face PAD models. Furthermore, the training, validation, and testing sets were split in subject-disjoint manner in the ECCV2020 challenge when the domain clues (\eg, recording sensors and lighting conditions) and PAIs are too similar. As a result, even with the off-the-shelf deep models with powerful representation learning capacity, the participants can easily train on ensemble of several overfitting models in these two competitions, performing well not only on training and validation sets but also on test data. It is necessary to mimic the requirements of real-world high security applications and to design and capture even more challenging and larger scale datasets with unknown domains and unseen PA types in the test sets. 

TPR@FPR was utilized in the CVPR2019 and ECCV2020 challenges as the ranking criteria, while ACER was adopted in the three remaining recent face PAD competitions. Despite being widely used in large-scale biometric evaluations (\eg, face recognition), TPR@FPR was first utilized in face PAD competitions due to emerging larger scale face PAD datasets (\eg, the CASIA-SURF~\cite{zhang2019dataset} with 295,000 frame samples and CelebA-Spoof~\cite{zhang2020celeba} with over 62,000 samples), which made the calculation of TPR@FPR=1e-4 or even TPR@FPR=1e-6 possible. In most of the competitions (4 out of 5), the results were reported also reported using the mainstream metrics ACPER, BPCER, and ACER recommended in ISO/IEC 30107-3 standard~\cite{ACER}. However, the selection of score threshold value for computing the ACER is worth discussing from both contest and database design point of views. For the CVPR2019~\cite{cvpr2019challenge}, CVPR2020~\cite{cvpr2020challenge} and ICCV2021~\cite{liu20213d} challenges, the ACERs for the test sets were determined by the EER operating point on the validation sets, while in the IJCB2021 LivDet-Face~\cite{purnapatra2021face}) evaluation, the fixed threshold of 50 (normalized output liveness scores from 0 to 100) was utilized directly and it was up to the participant how the scores were normalized within the valid range of liveness scores. The former approach seems to be more reasonable but with some drawbacks, For instance, significant discrepancies in performance can be observed among the ranked solutions when comparing their results using the ROC related TPR@FPR metric and ACER (\eg, the second best solution in terms of ACER ranking in Table~\ref{tab:single_results} performed poorly in terms of TPR@FPR=1e-3 in Fig.~\ref{fig:sroc}). Furthermore, in practical biometric applications, the suitable operating point depends highly on the application context, but when looking at a single ACER value, the misclassification rates between bona fide and attack classes are often ignored that is crucial piece of information. For instance, a method performing well in terms of ACER at the selected threshold might suffer from severely imbalanced APCER/BPCER ratio. The latter evaluation method with fixed range of liveness scores and ACER threshold sounds intuitive from interpretability and real-world applications point of views, and reasonable considering the lack of specific validation set.

The CVPR2019~\cite{cvpr2019challenge}, CVPR2020~\cite{cvpr2020challenge} and ICCV2021~\cite{liu20213d} challenges can be considered to more transparent and fair due to the richer amount of publicly available information about the participating teams and the best-performing solutions, whereas the ECCV2020~\cite{zhang2021celeba} and IJCB2021 LivDet-Face~\cite{purnapatra2021face} contests reported limited information about the teams and evaluated solutions. In general, competitions should encourage the participants to provide authentic public details on the registered teams and the open-source implementations, as well as the detailed ablation analysis of the best-performing solutions. The authentic team information is useful for avoiding malicious registration situations where teams consisting of similar members from the same institution could make more submission entries, thus less cost of trial and error. The open-source codes and detailed ablation analysis would benefit the reproducibility of the solutions and mitigate the possibility of cheating using manually annotated competition test data. The best solution to prevent "data peeking" would be to keep the evaluation set, including unseen test conditions, inaccessible during algorithm development phase and to conduct independent (third-party) evaluations, in which the organizers run the provided executables or source codes of the submitted systems on the competition data. At minimum, the organizers should be able to retrain and rerun the best-performing models following the official competition protocols to check if the submitted solutions have been calibrated, or even trained, on the test set and determine the final ranking of the teams based on the verified results.

The ICCV2021 challenge~\cite{liu20213d} provided also a fairer evaluation of the proposed algorithms per se by limiting the influence caused by differences in the amount of training data and number of ensembled (deep) models. The use of same training data and fixed number of models would be fairer for algorithm performance evaluation. Otherwise, the competitions just assess and ascertain how far the participants can push the face PAD performance with "black-box" methods on the specific benchmark, while not gaining actual insight in the effectiveness of the different proposed algorithms under the same conditions. For instance, we observed that the performance gains with the best solutions of the CVPR2019 \cite{cvpr2019challenge} and ICCV2021~\cite{liu20213d} challenges were largely due to the use of large-scale pretraining data and huge ensemble models, respectively. Another option would be to conduct separarate ablation studies on the competition test data by evaluating the solutions trained on the same training data in order to find out the impact of data, especially in the case of proprietary datasets.

The CASIA-SURF~\cite{zhang2020casia,zhang2019dataset} and the CASIA-SURF CeFA dataset~\cite{liu2021casia} used in the CVPR2019 \cite{cvpr2019challenge} and CVPR2020~\cite{cvpr2020challenge} challenges, respectively, provide pre-cropped and aligned facial images for each modality. Furthermore, the background information has been pixel-wise masked out to mitigate the effect of different face detection and alignment methods and limit the problem of face PAD to the actual facial information instead of exploiting the domain-specific contextual cues. The findings of the ICCV2021 challenge ICCV2021~\cite{liu20213d} suggest, however, that the use of proper pre-processing (\eg, face or facial attribute localization) can, in fact, significantly improve the PAD performance as the best-performing teams used additional pre-processing steps to focus on the actual discriminative facial details and specific sub-region information and to avoid learning and extraction of irrelevant features for face PAD. Data pre-processing needs definitely further attention in future studies because it is an understudied subject in face PAD research and an important component in complete face PAD solutions used in real-world AFR applications. While artificially restricting the original facial images and videos into pre-cropped bounding boxes or pixel-wise masked regions mitigates the issues related to sharing and working with huge datasets, and exploiting dataset-specific contextual cues for better PAD performance, the heavily pre-processed data can also limit the novelty of proposed solutions and usability of the dataset for experimental analysis during the competitions and, more importantly, in future studies in the research field. Therefore, the research community needs to find means for providing large-scale face PAD datasets also with unprocessed facial images and videos in order to be able to evaluate complete face PAD solutions and conduct comprehensive ablation studies.

\subsection{Future challenges}

The test cases in the current competitions measuring the generalization across the different covariates are still rather limited. Especially, the domain diversity should be increased as the samples in most (4 out of 5) of the competitions were recorded in indoor office locations with no more than three ethnicities. Regarding the PA  species, recently introduced challenging partial face attacks (\eg, half 3D mask, makeup and tattoo) have not been yet considered in face PAD contests. Another one issue is the imbalanced long-tailed data distribution across different PAs, where some more challenging attack types (\eg, high-fidelity masks) have been represented usually only with a few samples due to their high manufacturing costs. A large-scale test set 'in the wild' with diverse domain conditions and PA types, as well as more balanced data distribution will be eventually needed to achieve more realistic evaluation settings.  

Most of the existing face PAD competitions have not considered the efficiency or costs of the proposed solutions as no constraints on either model size and number of models have been given. As a result, the best-performing algorithms have been usually ensembles of several deep models, which gives insight how robust PAD performance can be reached with the current state-of-the-art methods on the competition data. However, blindly pushing to maximum detection performance without any restrictions encourages the participants to exploit and combine off-the-shelf components instead of trying to invent truly novel and effective solutions that could be actually deployed in real-world applications on mobile and embedded platforms with restricted resources. Although the organizers of the latest ICCV2021 challenge~\cite{liu20213d} explicitly informed that results obtained with fusion of deep models are rejected and the computational cost of a single model should be less than 100G FLOPs, there are still no evaluation metrics for measuring trade-off between accuracy and efficiency. 

Despite two multi-modal face PAD competitions have been already conducted, it is a bit disappointing that only few solutions pursued to introduce new kinds of advanced multi-modal fusion algorithms. Many major manufacturers have already included multi-modal camera systems into their products, including mobile devices and laptops, thus there is an urgent need to explore novel multi-modal fusion algorithms instead of just ensembling models with different modalities. Furthermore, not all the modalities are available as the selection of multi-modal data sources depends on the deployment scenario in question. Thus, it would be useful to investigate performance in setting where a model is trained on multiple modalities but evaluated on partial or arbitrary combinations of modalities. Among emerging imaging technologies, depth and NIR cameras have been already considered in the two multi-modal competitions, thus it would be interesting include also more advanced sensors, such as short-wave infrared (SWIR)~\cite{heusch2020deep}) or even hyperspectral imaging \cite{kaichi2021}, in upcoming collective evaluations. 

Apart from conventional PAs, two kinds of physical adversarial attacks (\ie, recognition and PA aware) could be considered for generic face PAD. For example, special printed eyeglasses~\cite{sharif2019general}, hats~\cite{komkov2021advhat} and stickers~\cite{guo2021meaningful} synthesized by adversarial generators have been demonstrated to effectively fool deep learning based AFR systems when worn by an attacker. Besides recognition-aware adversarial attacks, adversarial print and replay attacks~\cite{zhang2019attacking} with specific perturbation injected before physical broadcast have been developed to fool face PAD systems. Therefore, it can be expected to be necessary to consider diverse set of physical adversarial attacks in future competitions. In addition to PAs, there are many vicious digital manipulation attacks (\eg, "deepfakes"~\cite{rossler2019faceforensics}) that can be applied against AFR systems. Despite the differences in generation techniques and visual quality, some of these attacks still have coherent properties and artefacts. In~\cite{deb2021unified}, a unified digital and physical face attack detection framework is proposed to learn joint representations for coherent attacks. Therefore, another interesting challenge to tackle in upcoming contests assessing the robustness of face biometric systems would be to simultaneously detect both digital and physical attacks.

\section{Conclusions}
\label{sec:conclusion}

Competitions play a vital role in consolidating the recent trends and assessing the state of the art in face PAD. This chapter introduced the design and results of the five latest international competitions on unimodal and multi-modal face PAD organized from 2019. These contests have been important milestones in advancing the research on face PAD to the next level as each competition has offered new challenges to the research community and resulted in novel countermeasures and new insight. The industrial participants have dominated most of these competitions, which indicates the strong need for robust anti-spoofing solutions in real-world applications. The first two face PAD competitions (\ie, CVPR2019 \cite{cvpr2019challenge} and CVPR2020 \cite{cvpr2020challenge}) provided initial assessments of the state of the art in multi-modal face PAD algorithms under mainstream PAs in diverse conditions, while the three most recent face PAD competitions (\ie, ECCV2020 \cite{zhang2021celeba}, IJCB2021 LivDet-Face \cite{purnapatra2021face} and ICCV2021 \cite{liu20213d}) benchmarked conventional colour (RGB) camera based PAD algorithms under larger scale datasets, challenging unseen attacks and high-quality 3D mask attacks. Although several solutions proposed in the first two multi-modal competitions achieved satisfying PAD performance, more comprehensive multi-modal datasets and evaluation protocols on generalized PAD are still needed, especially considering the situation when missing a modality (or modalities) missing in training or test phase. In contrast, none of the systems proposed in the context of the latest two unimodal competitions managed to achieve satisfying PAD performance under unseen attacks detection in unknown operating conditions, thus more diverse larger scale unimodal face PAD datasets are still needed to develop and evaluate more robust learning based algorithms. 

\begin{acknowledgement}
This work was supported by the Academy of Finland for Academy Professor project EmotionAI, ICT 2023 project and Infotech Oulu.
\end{acknowledgement}

\ifthenelse{\equal{false}{\buildbook}}{
\printindex
\printglossary
\bibliographystyle{spmpsci}
\bibliography{references}
}

\end{document}